\documentclass{Interspeech}
\usepackage{newtxtext}

\interspeechcameraready

\author[affiliation={1,2}]{Vaibhav}{Rathore}
\author[affiliation={1,2}]{Siddhant}{Gole}
\author[affiliation={1}]{Dadhichi}{Telwadkar}
\author[affiliation={1}]{Rooshil}{Bhatia}
\author[affiliation={1}]{Maulik}{Ruparel}
\author[affiliation={1}]{Siddharth}{Sureka}
\author[affiliation={1}]{Neha}{Bhargava}

\usepackage[utf8]{inputenc}
\usepackage{ulem}
\usepackage{xcolor}
\usepackage{hyperref}
\usepackage{url}            
\usepackage{booktabs}       
\usepackage{amsfonts}
\usepackage{nicefrac}       
\usepackage{microtype}      
\usepackage{graphicx}       
\usepackage{amsmath} 
\usepackage{algorithm}
\usepackage{algpseudocode}
\usepackage[table]{xcolor}
\usepackage{comment}
\usepackage{amsmath} 
\usepackage{tcolorbox}
\usepackage{multirow}
\usepackage{tikz}
\usepackage{booktabs}
\usetikzlibrary{shapes.geometric, arrows.meta, positioning, calc, decorations.pathreplacing, backgrounds, fit}
\usepackage{caption}
\definecolor{failred}{HTML}{D32F2F}
\definecolor{successgreen}{HTML}{388E3C}
\definecolor{neutralgray}{HTML}{616161}

\affiliation{1}{Motilal Oswal Financial Services Ltd.}{India}
\affiliation{2}{Indian Institute of Technology (IIT) Bombay}{India}
\email{\{rathore.vaibhav, siddhant.gole, dadhichi.telwadkar, rooshil.bhatia, maulik.ruparel, siddharth.s, neha.bhargava\}@motilaloswal.com}


\title{\textsc{SuTRA} : Structurally-Unified Tokenization with Root Awareness}

\begin{document}

\maketitle

\begin{abstract}
Existing subword tokenizers optimize statistical compression but ignore morphological structure, particularly the relationship between roots and affixes. This is harmful for morphologically rich Indic languages, where basic units are complex orthographic syllables (\textit{aksharas}) rather than letters. Frequency-based methods over-fragment words, arbitrarily splitting roots and affixes—a phenomenon we term \textit{Morphological Shattering}. We propose \textsc{SuTRA}\footnote{Project Page: \href{https://mo-vaibhavr-43300.github.io/SuTRA/}{https://mo-vaibhavr-43300.github.io/SuTRA/}} (Structurally-Unified Tokenization with Root Awareness), a morphology-aware algorithm that preserves \textit{akshara} indivisibility and penalizes merges crossing morphological boundaries. We also release a new morphological segmentation dataset for Hindi, Marathi, and Gujarati. \textsc{SuTRA} reduces shattering, achieving peak gains of +14.7\% in morphological alignment (Boundary F1) and +34\% in semantic recoverability (Hindi) over \textsc{Bpe}. These structural gains yield an average improvement of +8.08 \textsc{chrF2} in machine translation. 
\end{abstract}

\section{Introduction}

Modern \textsc{NLP} pipelines rely on tokenizers as the foundational bridge to resolve the input-representation gap, converting raw text into the discrete tokens that populate the high-dimensional vector spaces of Language Models \cite{pennington2014glove, mikolov2013distributed, sarzynska2021detecting}. Current \textit{de facto} standards---such as Byte-Pair Encoding (\textsc{Bpe}) \cite{Gage1994-ds, Sennrich2016-uh}, \textsc{WordPiece} \cite{Schuster2012-ev}, and \textsc{Unigram} \cite{Kudo2018-ue}---function primarily as statistical data compression techniques. While they efficiently reduce sequence lengths and out-of-vocabulary (OOV) rates, they remain fundamentally agnostic to the morphological structure of language \cite{Hofmann2022-td, Hofmann2020-lg}. Despite explorations into tokenization-free architectures \cite{deiseroth2024t, clark2022canine, xue2022byt5, tay2021charformer}, state-of-the-art models \cite{brown2020language, guo2025deepseek, yang2025qwen3, team2025gemma} continue to rely on \textsc{Bpe}-style segmentation and inherit its structural limitations.

\begin{figure}[t]
    \centering
    \includegraphics[width=\linewidth]{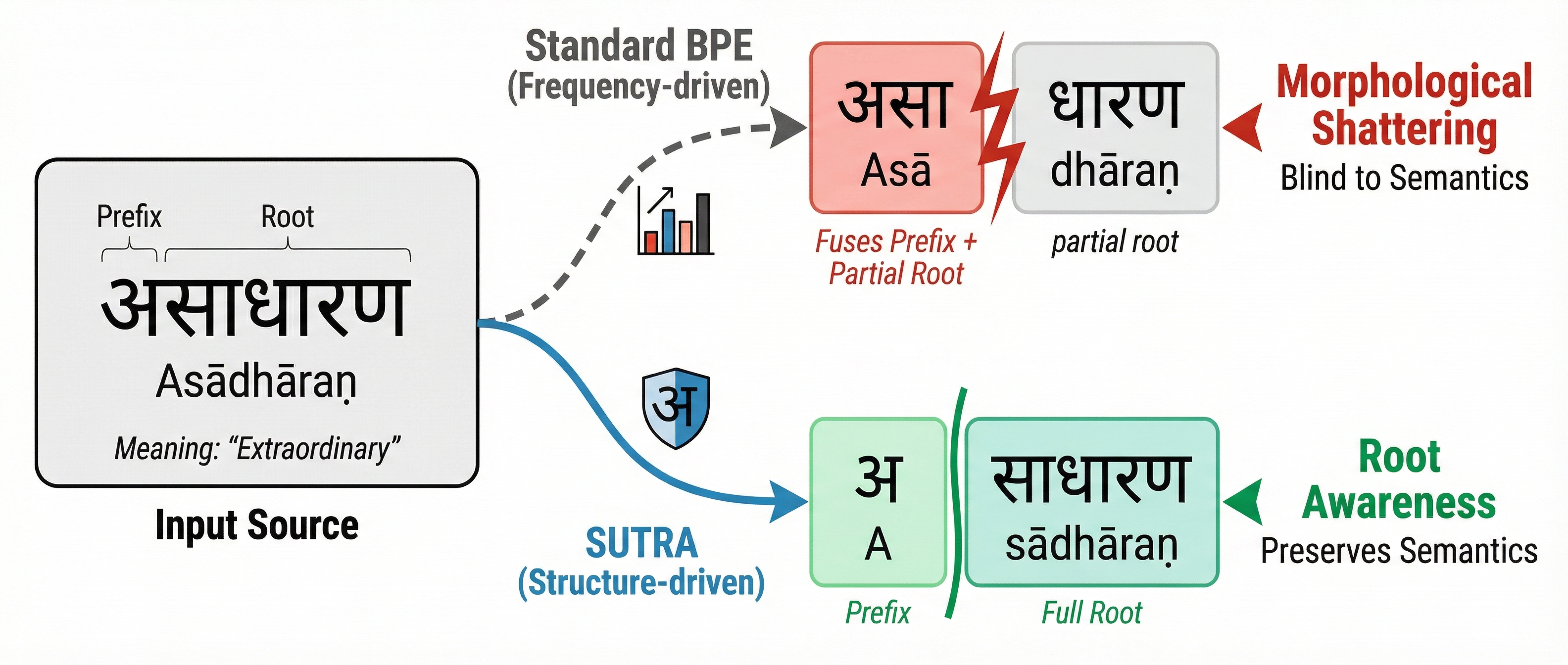}
    \vspace{-1.0em}
    \caption{\textbf{Morphological Shattering vs.\ Root Preservation.} For the Hindi word \textit{asādhāraṇ}, standard \textsc{Bpe} fuses the negation prefix into the root (\texttt{[asā]+[dhāraṇ]}), while \textsc{SuTRA} cleanly separates prefix and root (\texttt{[a]+[sādhāraṇ]}). We term such frequency-driven prefix--root fusion \textit{Morphological Shattering}; \textsc{SuTRA}'s root-preserving segmentations yield more stable subword units and reduce semantic blindness. Figure generated using PaperBanana~\cite{paperbanana}.}
    \label{fig:teaser}
    \vspace{-3.0em}
\end{figure}

Because these tokenizers ignore morphological rules, they often segment multimorphemic words into sub-tokens that misalign with lexical roots and affixes. We refer to this tokenizer-level misalignment as \textit{Morphological Shattering} (Figure~\ref{fig:teaser}). The problem is amplified in morphologically rich languages, especially those written in Indic scripts such as Devanagari. These scripts are abugidas: base consonants and dependent vowels (\textit{matras}) together form orthographic syllables (\textit{aksharas}) that function as atomic written units~\cite{akshara_ref}. Off-the-shelf tokenizers frequently split inside these units---for example, separating a \textit{matra} from its base consonant. Many Indo-Aryan languages also exhibit \textit{Sandhi}, where segments at word and morpheme boundaries change or fuse in the surface form \cite{sandhan2022translist,gaikwad2021state}, further obscuring the underlying morpheme boundaries. Together with the well-documented \textit{Indic Tax} of higher token fertility (more tokens per word) \cite{kumar2026sanskrit,pattnayak2025tokenization,tamang2024evaluating,chaudhari2023significance}, these properties lead to more severe Morphological Shattering. We use \textit{Semantic Blindness} for the corresponding representation-level effect: root semantics become harder to recover with simple linear probes, and embeddings become overly sensitive to small orthographic changes that leave the underlying morphemes intact \cite{asgari2025morphbpe,arnett2025evaluating,morphtok2024,isac2025slip}.

To address Morphological Shattering and its semantic consequences in complex languages, we make three contributions:
\begin{itemize}
    \item \textbf{\textsc{SuTRA} (Structurally-Unified Tokenization with Root Awareness)}: We introduce a structure-guided extension of \textsc{Bpe} that augments frequency-based merging with lightweight linguistic priors. It combines script-aware grouping of akshara-like units with morphology-aligned merge scoring and a root-preserving boundary constraint to penalize splits across valid morphological units.
    
    \item \textbf{Indic Morphological Gold-Standard Dataset}: We construct a large-scale, LLM-assisted morphological segmentation dataset tailored for three Indic languages (Hindi, Marathi, and Gujarati), providing a robust common benchmark for evaluating morphological alignment.
    
    \item \textbf{Intrinsic and Downstream Evaluation}: We design a comprehensive evaluation suite spanning morphological alignment, semantic recoverability, robustness under orthographic perturbations, and machine translation. Across these four axes, \textsc{SuTRA} reduces Morphological Shattering and yields more stable, structurally aligned representations: it better respects morphemic boundaries without arbitrary over-segmentation and improves both intrinsic semantic recoverability and downstream quality compared to standard baselines.
\end{itemize}

\section{Related Work}
\subsection{Frequency-Driven Tokenization}
Standard vocabulary construction relies on compression algorithms like \textsc{Bpe} \cite{Gage1994-ds, Sennrich2016-uh}, \textsc{WordPiece} \cite{Schuster2012-ev}, \textsc{Unigram} \cite{Kudo2018-ue}, and \textsc{SuperBpe} \cite{liu2025superbpespacetravellanguage}. By treating text as a language-agnostic character stream, these methods maximize statistical coverage but frequently override linguistic boundaries, systematically provoking \textit{Morphological Shattering} \cite{Banerjee2018-hd, Hofmann2020-lg}. Prior mitigations, such as stochastic regularization (\textsc{Bpe-Dropout} \cite{provilkov-etal-2020-bpe}) or post-hoc realignment (\textsc{Flota} \cite{hofmann-etal-2022-embarrassingly}), remain auxiliary interventions that fail to correct the structural dissociation ingrained during initial vocabulary creation. Furthermore, standard evaluation metrics like Compression Ratio and Perplexity \cite{haga2025babylmchallengeexploringeffect, ali2024tokenizer} mask these representational inefficiencies because they measure sequence predictability rather than sub-lexical integrity. 

\subsection{Character-level and Token-free Architectures}To bypass the limitations of subword vocabularies, several works have proposed tokenization-free or character-level architectures such as \textsc{Canine} \cite{clark2022canine}, \textsc{ByT5} \cite{xue2022byt5}, and \textsc{Charformer} \cite{tay2021charformer}. These models operate directly on raw bytes or characters, theoretically avoiding the "shattering" problem entirely by allowing the model to learn its own internal representations of morphology \cite{deiseroth2024t}. However, these architectures suffer from a significant \textit{Compute Penalty}, as character-level processing drastically increases sequence lengths, making them computationally expensive for long-context tasks. Consequently, state-of-the-art models continue to utilize subword-based \textsc{Bpe} \cite{guo2025deepseek, yang2025qwen3, team2025gemma}, inheriting its structural limitations while benefiting from its efficiency. 

\subsection{Morphological Challenges in Indic \textsc{LLM}s}\textit{Morphologically Rich Languages} (\textsc{Mrl}s), particularly those in the \textit{Indic} family, suffer from a disproportionate \textbf{Indic Tax} characterized by high fertility rates and poor semantic anchoring \cite{morphtok2024, Lian2025-ch}. Traditional \textsc{Bpe} fails to capture \textit{Sandhi} (phonetic fusions) and routinely violates the integrity of the \textit{Abugida} script by decoupling \textit{matras} (dependent vowels) from base consonants \cite{Banerjee2018-hd}. Prior work has explored morphology-aware tokenization and attention-guided segmentation \cite{morphtok2024, asgari2025morphbpe, jabbar2024morphpiecelinguistictokenizer}, but either treats morphology as a pre-tokenization heuristic or targets specific languages (English, Hebrew, Turkish) rather than Indic abugidas \cite{jabbar2024morphpiecelinguistictokenizer, seker-etal-2022-alephbert, Toraman_2023}. Although contemporary approaches like \textsc{Ag-Bpe} \cite{charlet_2025_agbpe_v3} target semantic awareness through attention-guided scoring, it fails to explicitly preserve the atomic boundaries of Indic scripts. In contrast, our approach works with penalty driven morphological violations, which exhibit higher sensitivity to preserving token-level semantic coherence.



\section{Morphological Dataset Preparation}
\label{sec:dataset_construction}
A critical bottleneck for \textit{MRL} subword tokenization is the lack of verified ground-truth segmentation. Large-scale resources like Unimorph 4.0~\cite{batsuren2022unimorph40universalmorphology} and MorphyNet~\cite{batsuren-etal-2021-morphynet} are primarily \textit{paradigm-centric}—mapping lemmas to inflected forms without defining explicit boundaries. Similarly, specific datasets like GujMORPH~\cite{baxi-bhatt-2022-gujmorph} rely on rule-based stemming, failing to restore oblique roots or handle phonetic fusion (\textit{Sandhi}). To address this, we constructed a \textbf{Gold Standard Morphological Split Dataset} for \textit{Hindi}, \textit{Marathi}, and \textit{Gujarati}. As shown in Table~\ref{tab:dataset_comparison}, our dataset enforces \textbf{Root Restoration} (e.g., mapping oblique \textit{ghara} to root \textit{ghar}) via an LLM-in-the-loop pipeline.

\begin{table}[ht]
\centering
\small
\setlength{\tabcolsep}{3pt}

\resizebox{\columnwidth}{!}{%
\begin{tabular}{l l r c l}
\toprule
\textbf{Dataset} & \textbf{Langs (Indic)} & \textbf{Size} & \textbf{Method} & \textbf{Verification} \\
\midrule
UniMorph 4.0 \cite{batsuren2022unimorph40universalmorphology} & Multi (Partial) & $\sim$10M & Schema-based & Algorithmic \\
MorphyNet \cite{batsuren-etal-2021-morphynet} & 15 (\textbf{None}) & 10.6M & Rule+Enrichment & Manual (Expert) \\
GujMorph \cite{baxi-bhatt-2022-gujmorph} & 1 (Gujarati) & $\sim$80k & Unsup. & None \\
\midrule
\textbf{Ours (Gold)} & \textbf{3(Hi, Mr, Gu)} & \textbf{560k} & \textbf{Hybrid} & \textbf{\textsc{LLM}-Verif.} \\
\bottomrule
\end{tabular}
}
\caption{\textbf{Morphological Datasets.} We provide the first high-scale, LLM-verified morphological coverage for Indic scripts.}
\vspace{-1.5em}
\label{tab:dataset_comparison}
\end{table}

\subsection{Dataset Pipeline}

We employ a three-stage hybrid pipeline—corpus extraction, unsupervised decomposition, and LLM verification (details in \textbf{Sup. Mat.}). First, we extract a diverse vocabulary from \textit{IndicCorp}~\cite{ai4bharat_corpus}. For decomposition, we adapt \textsc{SampoNlp}~\cite{chelombitko2025samponlp} to the Indo-Aryan context, leveraging its \textsc{Minimum Description Length} atomicity scoring while modifying degeneracy constraints to accommodate single-character Indic affixes and \textit{matras}. Next, \textsc{Gemini 3.0 Flash}~\cite{team2023gemini} resolves ambiguities (e.g., \textit{Sandhi}) and corrects over-segmentation. This yields a Gold Standard lexicon of $\sim$\textbf{560,000} verified words (Table~\ref{tab:dataset_stats}), acting as \textsc{SuTRA}'s morphological probe. Crucially, we employ a \textbf{dual-layer logic}: the lexicon stores pure \textit{canonical splits}, while \textsc{SuTRA} maps these to \textit{surface boundary indices} during training. We prioritize \textit{surface-form integrity} over canonical purity to guarantee zero-overhead detokenization, enabling exact recovery via simple concatenation.


\begin{table}[ht]
\centering
\small
\begin{tabular}{l l r}
\toprule
\textbf{Language} & \textbf{Source} & \textbf{Unique Words} \\
\midrule
\textit{Hindi} & \textit{IndicCorp}~\cite{ai4bharat_corpus} & $\simeq$ 160,000 \\
\textit{Marathi} & \textit{IndicCorp}~\cite{ai4bharat_corpus} & $\simeq$ 200,000 \\
\textit{Gujarati} & \textit{IndicCorp}~\cite{ai4bharat_corpus} & $\simeq$ 200,000 \\
\midrule
\textbf{Total} & & $\simeq$ \textbf{560,000} \\
\bottomrule
\end{tabular}
\caption{\textbf{Gold Standard Statistics.} Hybrid of rule-based extraction and \textsc{LLM}-verified morphological segmentation.}
\vspace{-1.0em}
\label{tab:dataset_stats}
\end{table}

\section{Methodology}

\begin{figure*}[!th]
    \centering
    \includegraphics[width=0.85\linewidth]{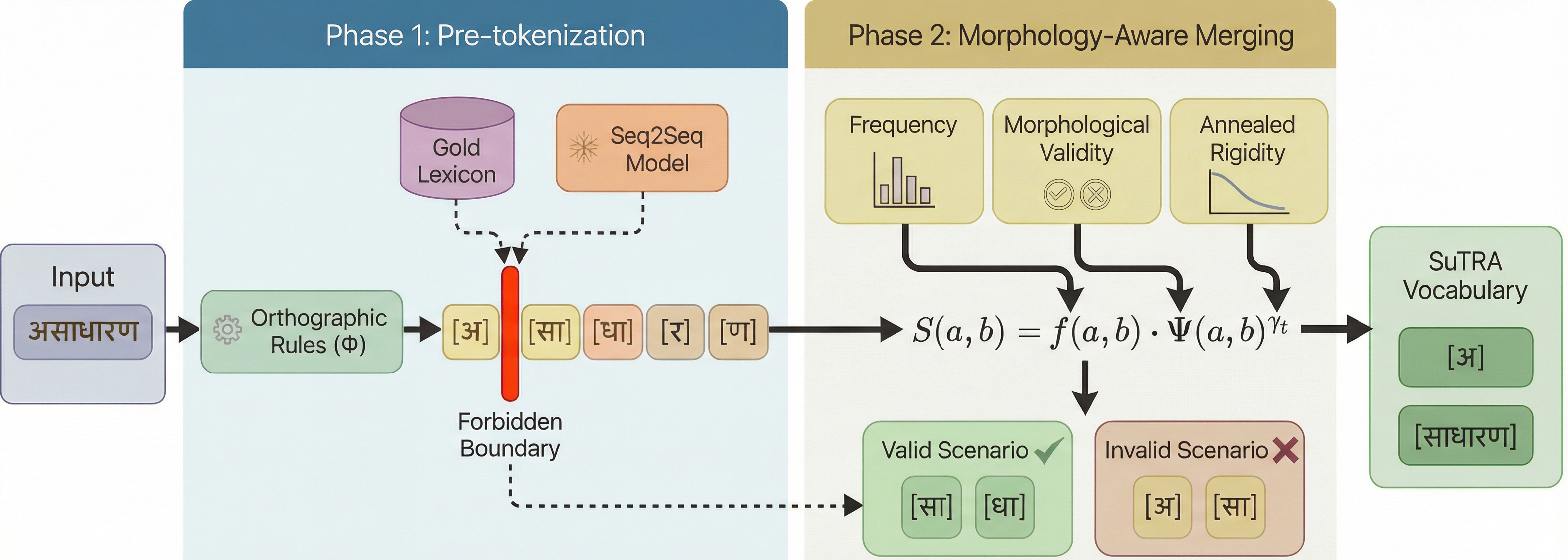}
    \caption{Overview of \textsc{SuTRA}. \textbf{Phase 1 (Pre-tokenization)} applies orthographic rules $\Phi$ to map each word into akshara-like units and uses a gold morphological lexicon or a seq2seq model to mark \textit{forbidden boundaries} (morpheme boundaries that merges should not cross). \textbf{Phase 2 (Morphology-Aware Merging)} runs a BPE-style algorithm with scores $S(a,b) = f(a,b)\,\Psi(a,b)^{\gamma_t}$, where $\Psi$ downweights merges that violate forbidden boundaries and $\gamma_t$ controls rigidity over training, biasing vocabulary toward merges that respect Indic script structure and morpheme boundaries. Figure generated using PaperBanana~\cite{paperbanana}.}
    \label{fig:scbpe_workflow}
    \vspace{-1.2em}
\end{figure*}

We introduce \textsc{SuTRA}, a two-phase framework that integrates morphological priors into statistical subword learning (Figure \ref{fig:scbpe_workflow}). The pipeline systematically transitions tokenization from purely frequency-driven to morphologically grounded segmentation.

\noindent\textbf{Phase 1: Pre-tokenization.}
Building on \textsc{MorphTok}, this phase defines atomic units and marks morphologically sensitive boundaries. Standard tokenizers often isolate dependent vowels (\textit{matras}), breaking the akshara structure of Indic abugida scripts. To avoid such script-breaking splits, we apply orthographic rules ($\Phi$) that map each word $w$ to a sequence of akshara-like units $U = [u_1, u_2, \dots, u_n]$, binding modifiers to their base consonants. In parallel, we perform a morphological lookup to flag \textit{forbidden} boundaries—boundaries between akshara units that coincide with morpheme boundaries in our gold lexicon. A curated lexicon is used for known words, and a fine-tuned character-level seq2seq model infers boundaries for out-of-vocabulary items.

\begin{figure}[ht]
    \centering
    \includegraphics[width=0.85\linewidth]{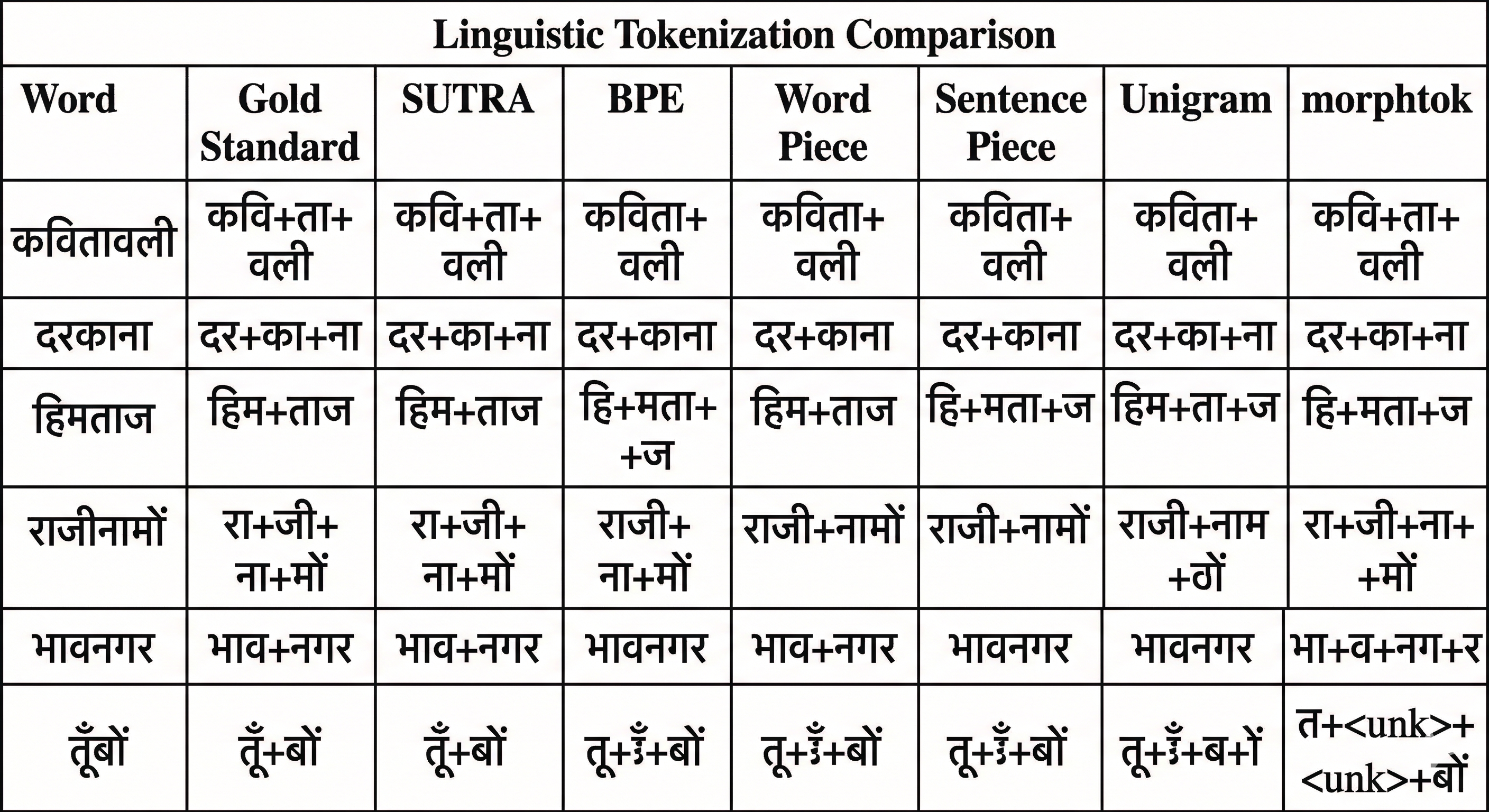}
    \caption{Qualitative Comparison of Morphological Segmentation Across Tokenizers. \textsc{SuTRA} consistently matches the Gold Standard by respecting phonetic and morphological boundaries.}
    \label{fig:comaparison_table}
    \vspace{-1.5em}
\end{figure}

\noindent\textbf{Phase 2: Training (Morphology Aware Merging).} While \textsc{MorphTok} relies on rigid, pre-computed boundary enforcement, \textsc{SuTRA} differentiates itself by integrating these morphological priors directly into a score-based \textsc{Bpe} merging process. Instead of pure frequency-based merging, \textsc{SuTRA} dynamically penalizes candidate merges that cross the forbidden boundaries identified in Phase 1. Let $f(a,b)$ be the corpus frequency of a candidate pair $(a,b)$, and $\chi(a,b)$ be its \textit{conflict count} (the number of occurrences where merging $(a,b)$ crosses a forbidden boundary). The morphological validity probability $\Psi(a,b) \in [0,1]$ is:
$$ \Psi(a,b) = 1 - \frac{\chi(a,b)}{f(a,b)} $$
To govern the merging process, we introduce a dynamic \textbf{Rigidity Constraint} ($\gamma_t$). The final merge score $S(a,b)$ is defined as:
$$ S(a,b) = f(a,b) \cdot \Psi(a,b)^{\gamma_t} $$
During training, $\gamma_t$ is annealed from $\gamma_{start}$ to $\gamma_{end}$. This exponential decay forms a curriculum: the algorithm strictly prioritizes merging safe, continuous lexical roots early in training (high rigidity) before relaxing the constraint to attach functional affixes later (low rigidity). Figure~\ref{fig:comaparison_table} shows that \textsc{SuTRA} constraints translate into cleaner segmentations that preserve the akshara structure and morpheme boundaries as compared to other tokenizers. The detailed algorithm along with complexity analysis is given in \textbf{Sup. Mat.}

\section{Experiments and Results}
To evaluate the efficacy of \textsc{SuTRA}, we conduct experiments on three morphologically rich Indic languages: Hindi (HI), Marathi (MR), and Gujarati (GU). Our evaluation is organized around four questions that mirror our central claims: (1) does \textsc{SuTRA} better align subwords with gold morpheme boundaries? (2) do these boundaries make whole-word semantics more directly recoverable from subword embeddings? (3) do the resulting representations yield measurable gains on downstream machine translation? and (4) does \textsc{SuTRA} maintain morphological robustness under orthographic perturbations? 
We therefore assess \textsc{SuTRA} along four complementary axes: morphological alignment, semantic recoverability, translation quality, and robustness to surface noise.

\subsection{Morphological Alignment Evaluation}
\label{sec:morph_eval}
We first evaluate whether tokenizer boundaries align with gold morpheme segmentations in Hindi (HI), Marathi (MR), and Gujarati (GU).
\noindent\textbf{Setup and Metrics.}
We normalize tokenizer outputs by stripping subword artifacts (e.g., \texttt{\#\#}, \texttt{\_}) to enable direct string matching with gold morphemes. We report: (1) \textbf{Boundary F1}—the harmonic mean of precision and recall over predicted vs.\ gold boundary indices; and (2) \textbf{Fertility Ratio}—the mean number of predicted tokens per gold segment ($|T_{\text{pred}}|/|T_{\text{gold}}|$).

\begin{table}[ht!]
\centering
\small
\resizebox{\columnwidth}{!}{%
\begin{tabular}{l|cc|cc|cc}
\hline
\textbf{Tokenizer} & \multicolumn{2}{c|}{\textbf{Hindi}} & \multicolumn{2}{c|}{\textbf{Marathi}} & \multicolumn{2}{c}{\textbf{Gujarati}} \\
\cline{2-7}
& \textbf{F1} $\uparrow$ & \textbf{Fert.} $\downarrow$ & \textbf{F1} $\uparrow$ & \textbf{Fert.} $\downarrow$ & \textbf{F1} $\uparrow$ & \textbf{Fert.} $\downarrow$ \\
\hline
BPE (ACL'16) & 0.482 & 1.285 & 0.470 & 1.225 & 0.591 & 1.126 \\
WordPiece (arXiv'12) & 0.411 & 1.214 & 0.527 & 1.300 & 0.596 & 1.173 \\
SentencePiece (EMNLP'18) & 0.438 & 1.315 & 0.083 & 1.183 & 0.591 & 1.156 \\
Unigram (ACL'18) & 0.439 & 1.310 & 0.507 & 1.256 & \textbf{0.669} & 1.137 \\
SuperBPE (COLM'25) & 0.089 & 2.517 & 0.084 & 3.084 & 0.096 & 3.113 \\
MorphTok (ICML-W'25)     & 0.190 & 2.601 & 0.216 & 3.482 & 0.247 & 3.494 \\
\midrule
\rowcolor[gray]{0.9} \textsc{SuTRA} (Ours) & \textbf{0.586} & 1.412 & \textbf{0.617} & 1.755 & 0.584 & 1.454 \\
\hline
\end{tabular}
}
\caption{\textbf{Morphological Alignment Evaluation.} Evaluation of Boundary F1 and Fertility Ratio across Hindi, Marathi, and Gujarati. Higher F1 indicates superior structural integrity, while controlled fertility prevents arbitrary character-level fragmentation. Best F1 scores are in bold.}
\label{tab:tok_results_clean}
\end{table}

\noindent\textbf{Results and Discussion.}
Table~\ref{tab:tok_results_clean} shows that \textsc{SuTRA} achieves the highest Boundary F1 for Hindi (0.586) and Marathi (0.617), and remains competitive for Gujarati, while keeping fertility within a moderate range. High-fertility baselines such as SuperBPE and \textsc{MorphTok} produce many more tokens per morpheme yet still underperform on F1, indicating over-fragmentation rather than genuine morphological alignment. By enforcing \textit{akshara} atomicity and penalizing merges that cross morpheme boundaries, \textsc{SuTRA}'s extra segments tend to correspond to functional affixes rather than the Morphological Shattering observed in purely frequency-driven tokenizers. Extended metrics are reported in the \textbf{Sup.\ Mat.}

\subsection{Semantic Recoverability}
Arbitrary subword fragmentation forces models to expend capacity reconstructing basic lexical semantics. We therefore ask whether \textsc{SuTRA}'s boundaries make whole-word meaning more directly recoverable from subword embeddings.

\noindent\textbf{Setup \& Metrics.} We train a Word2Vec model \cite{mikolov2013efficientestimationwordrepresentations} on a joint vocabulary containing both words and subwords. After training, we freeze these embeddings and train lightweight diagnostic models to reconstruct a target whole-word embedding from the mean-pooled embeddings of its constituent subwords. 
We measure \textit{semantic recoverability} using $R^2$, the proportion of variance in the target embedding that is predictable from the subwords. A \textbf{Linear} probe (no hidden layers) tests simple additive composition, while a \textbf{Multi-Layer Perceptron (MLP)} probe tests whether more complex morphological interactions remain recoverable (See \textbf{Sup. Mat.} for details.)

\begin{table}[ht]
\centering
\tiny
\resizebox{\columnwidth}{!}{%
\begin{tabular}{l|cc|cc}
\toprule
\multirow{2}{*}{\textbf{Language}} & \multicolumn{2}{c|}{\textbf{Linear $R^2$ (Layer 0)}} & \multicolumn{2}{c}{\textbf{MLP $R^2$ (Layer 2+)}} \\ 
\cmidrule(lr){2-3} \cmidrule(lr){4-5}
 & \textbf{\textsc{SuTRA}} & \textsc{Bpe} & \textbf{\textsc{SuTRA}} & \textsc{Bpe} \\ \midrule
\textit{Hindi} & \textbf{0.4464} & 0.3329 & \textbf{0.5048} & 0.3358 \\
\textit{Marathi} & \textbf{0.4634} & 0.4619 & \textbf{0.5331} & 0.4649 \\
\textit{Gujarati} & 0.4624 & \textbf{0.4640} & \textbf{0.5055} & 0.4510 \\
\bottomrule
\end{tabular}%
}
\caption{\textbf{Semantic Recoverability ($R^2$).} Linear models evaluate immediate compositionality, while deeper MLP models test the preservation of recoverable structural signals.}
\vspace{-1.0em}
\label{tab:linear_recover}
\end{table}

\noindent\textbf{Results \& Discussion.}
Table~\ref{tab:linear_recover} shows that \textsc{SuTRA} generally preserves semantic information more effectively than \textsc{Bpe}. In Hindi, \textsc{SuTRA} yields a +34\% relative gain in Linear $R^2$ over \textsc{Bpe}, and the MLP provides only marginal improvement for \textsc{Bpe} (0.33 $\rightarrow$ 0.34), suggesting that little additional structure is recoverable beyond simple addition. In Marathi and Gujarati, Linear scores are similar, reflecting dense orthography and complex stem alternations, but \textsc{SuTRA} benefits substantially from the MLP ($R^2 > 0.50$ in both cases), whereas \textsc{Bpe} improves only modestly. This confirms \textsc{SuTRA} preserves recoverable structural signals that arbitrary fragmentation destroys.

\subsection{Machine Translation Performance}
\label{sec:mt_exp}
To assess downstream utility, we evaluate \textsc{SuTRA} on a \textit{Hindi}$\leftrightarrow$\textit{Marathi} translation task using a 3-layer Transformer \cite{vaswani2017attention} trained on the \textit{BhasaAnuvaad} corpus \cite{jain2024bhasaanuvaad}.

\noindent\textbf{Implementation.}
We use a standard Transformer encoder--decoder ($L=3, H=4, d_{\mathrm{ff}}=400$) implemented in \texttt{fairseq} \cite{ott2019fairseq}. All models are trained for 100k updates with a shared 32k vocabulary per tokenizer and \textit{IndicNLP} normalization, keeping architecture and optimization hyperparameters identical across baselines to isolate the impact of segmentation (see \textbf{Sup.\ Mat.} for full details). 

\noindent\textbf{Results.} Table~\ref{tab:mt_results} shows that \textsc{SuTRA} attains the best scores for \textit{Marathi}$\rightarrow$\textit{Hindi} and remains competitive with the strongest baseline in the reverse direction, indicating that improved morphological alignment and semantic recoverability translate into meaningful gains for neural MT. A secondary evaluation on causal language modeling, along with metric derivations and full hyperparameter settings, is provided in the \textbf{Sup.\ Mat.}; these tasks follow standard evaluation protocols in prior work \cite{morphtok2024}.
\begin{table}[ht]
\centering
\resizebox{\columnwidth}{!}{%
\begin{tabular}{l|cc|cc}
\toprule
\textbf{Tokenizer} 
& \multicolumn{2}{c|}{\textbf{Hindi $\rightarrow$ Marathi}} 
& \multicolumn{2}{c}{\textbf{Marathi $\rightarrow$ Hindi}} \\ 
\cmidrule(lr){2-3} \cmidrule(lr){4-5}
& \textbf{chrF2 $\uparrow$} & \textbf{COMET $\uparrow$} 
& \textbf{chrF2 $\uparrow$} & \textbf{COMET $\uparrow$} \\
\midrule

\rowcolor[gray]{0.95} \multicolumn{5}{l}{\textit{Statistical Baselines}} \\
\textsc{Bpe (ACL'16)} & 24.62 & 0.5054 & \uline{36.55} & \uline{0.6253} \\
\textsc{WordPiece (arXiv'12)} & \textbf{30.37} & \textbf{0.6093} & 27.18 & 0.5223 \\
\textsc{SentencePiece (EMNLP'18)} & 25.01 & 0.4964 & 28.10 & 0.5147 \\
\textsc{Unigram (ACL'18)} & 26.55 & 0.5157 & 29.19 & 0.5349 \\

\midrule
\rowcolor[gray]{0.95} \multicolumn{5}{l}{\textit{Morphological Baselines}} \\
\textsc{SuperBpe (COLM'25)} & 15.84 & 0.3047 & 16.27 & 0.3422 \\
\textsc{MorphTok (ICML-W'25)} & 26.75 & 0.5750 & 29.42 & 0.6007 \\

\midrule
\rowcolor[gray]{0.9} \textbf{\textsc{SuTRA} (Ours)} 
& \uline{29.96} & \uline{0.5752} 
& \textbf{38.84} & \textbf{0.6554} \\

\bottomrule
\end{tabular}
}
\caption{\textbf{Machine Translation.} \textsc{SuTRA} achieves the highest scores in \textit{Marathi} $\rightarrow$ \textit{Hindi} and remains comparable to the strongest baseline in the reverse task, demonstrating that root contiguity improves cross-lingual alignment.}
\label{tab:mt_results}
\vspace{-1.5em}
\end{table}

\subsection{Morphological Robustness Evaluation}
We evaluate tokenizer resilience to small surface changes (e.g., typos or suffix substitutions) that should not affect the underlying root but often trigger unstable segmentations. Using an adversarial suite of 10{,}000 words per language (HI, MR, GU) with synthetic orthographic and inflectional perturbations, we measure robustness via \textit{Jaccard Overlap} (Jac.\,$\uparrow$) between original and perturbed segmentations and \textit{Root-Affected Distance} (R.Aff.\,$\downarrow$), which tracks how often perturbations alter the subwords covering the lexical root. 

\noindent\textbf{Results \& Discussion.} As shown in Table~\ref{tab:baseline_comparison_robust}, purely statistical baselines (\textsc{Bpe}, \textsc{Unigram}) exhibit high R.Aff., indicating that minor variations frequently bleed into the core lexeme. In contrast, \textsc{SuTRA} attains near-zero R.Aff.\ while maintaining higher Jaccard overlap, showing that morphological priors effectively insulate roots from peripheral noise. This structural stability complements our semantic-recoverability and MT results, confirming that \textsc{SuTRA} maintains consistent tokenization under perturbations. Extended details are in the \textbf{Sup.\ Mat.}.

\begin{table}[htbp]
\centering
\small
\resizebox{\linewidth}{!}{%
\begin{tabular}{l|cc|cc|cc}
\toprule
\multirow{2}{*}{\textbf{Tokenizer}} & \multicolumn{2}{c|}{\textbf{Hindi (HI)}} & \multicolumn{2}{c|}{\textbf{Marathi (MR)}} & \multicolumn{2}{c}{\textbf{Gujarati (GU)}} \\
\cmidrule(lr){2-3} \cmidrule(lr){4-5} \cmidrule(lr){6-7}
 & \textbf{Jac.($\uparrow$)} & \textbf{R.Aff.($\downarrow$)} & \textbf{Jac.($\uparrow$)} & \textbf{R.Aff.($\downarrow$)} & \textbf{Jac.($\uparrow$)} & \textbf{R.Aff.($\downarrow$)} \\
\midrule
\textsc{Bpe} & 0.386 & 0.228 & 0.305 & 0.324 & 0.319 & 0.296 \\
\textsc{Unigram} & 0.363 & 0.232 & 0.294 & 0.340 & 0.320 & 0.308 \\
\textsc{WordPiece} & 0.413 & 0.249 & 0.293 & 0.355 & 0.314 & 0.316 \\
\textsc{SentencePiece} & 0.371 & 0.235 & 0.298 & 0.325 & 0.321 & 0.287 \\
\textsc{SuperBpe} & 0.624 & 0.146 & 0.655 & 0.127 & 0.631 & 0.126 \\
\textsc{MorphTok} & 0.801 & 0.067 & 0.823 & 0.055 & 0.792 & 0.051 \\
\midrule
\rowcolor[gray]{0.9} \textbf{\textsc{SuTRA} (Ours)} & \textbf{0.875} & \textbf{0.042} & \textbf{0.885} & \textbf{0.038} & \textbf{0.788} & \textbf{0.040} \\
\bottomrule
\end{tabular}%
}
\setlength{\tabcolsep}{2pt}
\caption{\textbf{Morphological Robustness Comparison.} Results show that \textsc{SuTRA} effectively mitigates root fragmentation.}
\vspace{-1.0em}
\label{tab:baseline_comparison_robust}
\end{table}



\section{Conclusion and Future Work}
Our results tell a consistent story. When tokenization is constrained to respect Indic script structure and morpheme boundaries, Morphological Shattering is reduced, and whole-word semantics become more linearly recoverable from subword embeddings. This, in turn, increases robustness and yields tangible gains on downstream translation, without increasing token fertility or vocabulary size.  \textsc{SuTRA} thus validates our core hypothesis: guiding frequency-based subword learning with lightweight morphological priors is a practical way to improve both interpretability and utility of \textsc{LLM} tokenizers for morphologically rich languages. Future work includes extending \textsc{SuTRA} to other morphologically rich languages, exploring unsupervised anchoring for low-resource scripts, and assessing its impact on additional text and speech tasks such as Text to Speech (TTS) and Automatic Speech Recognition (ASR).


\section{Acknowledgements}

This research was conducted at Motilal Oswal Financial Services Limited. The authors gratefully acknowledge the organization for providing the resources, infrastructure, and support that made this work possible.

\section{Use of Generative AI Disclosure}
In accordance with Interspeech 2026 guidelines, the authors disclose the following uses of Generative AI tools during the execution of this research and the preparation of this manuscript:
\begin{itemize}
    \item \textbf{Dataset Construction:} We utilized Gemini 3.0 Flash~\cite{team2023gemini} as an LLM-in-the-loop verifier within our dataset pipeline to resolve morphophonological ambiguities (e.g., \textit{Sandhi}) and correct algorithmic over-segmentation.
    \item \textbf{Visualizations:} Figure~\ref{fig:teaser} and Figure~\ref{fig:scbpe_workflow} was generated utilizing PaperBanana~\cite{paperbanana}.
    \item \textbf{Writing Assistance:} LLMs were utilized for minor language polishing, structural refinement, and LaTeX formatting during the drafting process. 
\end{itemize}
The human authors rigorously reviewed, modified, and validated all AI-assisted outputs and take full responsibility for the final content, scientific accuracy, and integrity of this work.



\bibliographystyle{IEEEtran}
\bibliography{main}

@inproceedings{vaswani2017attention,
 author = {Vaswani, Ashish and Shazeer, Noam and Parmar, Niki and Uszkoreit, Jakob and Jones, Llion and Gomez, Aidan N and Kaiser, \L ukasz and Polosukhin, Illia},
 booktitle = {Advances in Neural Information Processing Systems},
 editor = {I. Guyon and U. Von Luxburg and S. Bengio and H. Wallach and R. Fergus and S. Vishwanathan and R. Garnett},
 pages = {},
 publisher = {Curran Associates, Inc.},
 title = {Attention is All you Need},
 url = {https://proceedings.neurips.cc/paper_files/paper/2017/file/3f5ee243547dee91fbd053c1c4a845aa-Paper.pdf},
 volume = {30},
 year = {2017}
}

@article{akshara_ref,
    author = {Beesley, Kenneth R.},
    title = {A Computational Theory of Writing Systems},
    journal = {Computational Linguistics},
    volume = {27},
    number = {3},
    pages = {464-467},
    year = {2001},
    month = {09},
    issn = {0891-2017},
    doi = {10.1162/coli.2000.27.3.464},
    url = {https://doi.org/10.1162/coli.2000.27.3.464},
    eprint = {https://direct.mit.edu/coli/article-pdf/27/3/464/1797839/coli.2000.27.3.464.pdf},
}

@misc{paperbanana,
      title={PaperBanana: Automating Academic Illustration for AI Scientists}, 
      author={Dawei Zhu and Rui Meng and Yale Song and Xiyu Wei and Sujian Li and Tomas Pfister and Jinsung Yoon},
      year={2026},
      eprint={2601.23265},
      archivePrefix={arXiv},
      primaryClass={cs.CL},
      url={https://arxiv.org/abs/2601.23265}, 
}

@article{team2023gemini,
  title={Gemini: a family of highly capable multimodal models},
  author={Team, Gemini and Anil, Rohan and Borgeaud, Sebastian and Alayrac, Jean-Baptiste and Yu, Jiahui and Soricut, Radu and Schalkwyk, Johan and Dai, Andrew M and Hauth, Anja and Millican, Katie and others},
  journal={arXiv preprint arXiv:2312.11805},
  year={2023}
}

@inproceedings{ai4bharat_corpus,
    title = "Towards Leaving No {I}ndic Language Behind: Building Monolingual Corpora, Benchmark and Models for {I}ndic Languages",
    author = "Doddapaneni, Sumanth  and
      Aralikatte, Rahul  and
      Ramesh, Gowtham  and
      Goyal, Shreya  and
      Khapra, Mitesh M.  and
      Kunchukuttan, Anoop  and
      Kumar, Pratyush",
    editor = "Rogers, Anna  and
      Boyd-Graber, Jordan  and
      Okazaki, Naoaki",
    booktitle = "Proceedings of the 61st Annual Meeting of the Association for Computational Linguistics (Volume 1: Long Papers)",
    month = jul,
    year = "2023",
    address = "Toronto, Canada",
    publisher = "Association for Computational Linguistics",
    url = "https://aclanthology.org/2023.acl-long.693",
    doi = "10.18653/v1/2023.acl-long.693",
    pages = "12402--12426",

}

@ARTICLE{Gage1994-ds,
  title   = "A new algorithm for data compression",
  author  = "Gage, Philip",
  journal = "The C Users Journal",
  volume  =  12,
  pages   = "23--38",
  year    =  1994
}

@INPROCEEDINGS{Sennrich2016-uh,
  title      = "Neural machine translation of rare words with subword units",
  booktitle  = "Proceedings of the 54th Annual Meeting of the Association for
                Computational Linguistics (Volume 1: Long Papers)",
  author     = "Sennrich, Rico and Haddow, Barry and Birch, Alexandra",
  editor     = "Erk, Katrin and Smith, Noah A",
  publisher  = "Association for Computational Linguistics",
  pages      = "1715--1725",
  month      =  aug,
  year       =  2016,
  address    = "Stroudsburg, PA, USA",
  conference = "Proceedings of the 54th Annual Meeting of the Association for
                Computational Linguistics (Volume 1: Long Papers)",
  location   = "Berlin, Germany"
}

@INPROCEEDINGS{Kudo2018-ue,
  title      = "Subword regularization: Improving neural network translation
                models with multiple subword candidates",
  booktitle  = "Proceedings of the 56th Annual Meeting of the Association for
                Computational Linguistics (Volume 1: Long Papers)",
  author     = "Kudo, Taku",
  editor     = "Gurevych, Iryna and Miyao, Yusuke",
  publisher  = "Association for Computational Linguistics",
  pages      = "66--75",
  month      =  jul,
  year       =  2018,
  address    = "Stroudsburg, PA, USA",
  conference = "Proceedings of the 56th Annual Meeting of the Association for
                Computational Linguistics (Volume 1: Long Papers)",
  location   = "Melbourne, Australia"
}

@INCOLLECTION{Schuster2012-ev,
  title     = "Japanese and korean voice search",
  booktitle = "2012 {IEEE} international conference on acoustics, speech and
               signal processing ({ICASSP})",
  author    = "Schuster, M and Nakajima, K",
  publisher = "IEEE",
  pages     = "5149--5152",
  month     =  mar,
  year      =  2012
}

@INCOLLECTION{Banerjee2018-hd,
  title     = "Meaningless yet meaningful: Morphology grounded subword-level
               {NMT}",
  booktitle = "Proceedings of the second workshop on subword/character level
               models",
  author    = "Banerjee, T and Bhattacharyya, P",
  pages     = "55--60",
  month     =  jun,
  year      =  2018
}

@INCOLLECTION{Hofmann2020-lg,
  title     = "{DagoBERT}: Generating derivational morphology with a pretrained
               language model",
  booktitle = "Proceedings of the 2020 Conference on Empirical Methods in
               Natural Language Processing ({EMNLP})",
  author    = "Hofmann, V and Pierrehumbert, J and Sch{\"u}tze, H",
  pages     = "3848--3861",
  month     =  nov,
  year      =  2020
}

@INCOLLECTION{Hofmann2022-td,
  title     = "An embarrassingly simple method to mitigate undesirable
               properties of pretrained language model tokenizers",
  booktitle = "Proceedings of the 60th Annual Meeting of the Association for
               Computational Linguistics",
  author    = "Hofmann, V and Schuetze, H and Pierrehumbert, J",
  publisher = "Short Papers",
  volume    =  2,
  pages     = "385--393",
  month     =  may,
  year      =  2022
}

@inproceedings{provilkov-etal-2020-bpe,
    title = "{BPE}-Dropout: Simple and Effective Subword Regularization",
    author = "Provilkov, Ivan  and
      Emelianenko, Dmitrii  and
      Voita, Elena",
    editor = "Jurafsky, Dan  and
      Chai, Joyce  and
      Schluter, Natalie  and
      Tetreault, Joel",
    booktitle = "Proceedings of the 58th Annual Meeting of the Association for Computational Linguistics",
    month = jul,
    year = "2020",
    address = "Online",
    publisher = "Association for Computational Linguistics",
    url = "https://aclanthology.org/2020.acl-main.170/",
    doi = "10.18653/v1/2020.acl-main.170",
    pages = "1882--1892"
}

@inproceedings{hofmann-etal-2022-embarrassingly,
    title = "An Embarrassingly Simple Method to Mitigate Undesirable Properties of Pretrained Language Model Tokenizers",
    author = "Hofmann, Valentin  and
      Schuetze, Hinrich  and
      Pierrehumbert, Janet",
    editor = "Muresan, Smaranda  and
      Nakov, Preslav  and
      Villavicencio, Aline",
    booktitle = "Proceedings of the 60th Annual Meeting of the Association for Computational Linguistics (Volume 2: Short Papers)",
    month = may,
    year = "2022",
    address = "Dublin, Ireland",
    publisher = "Association for Computational Linguistics",
    url = "https://aclanthology.org/2022.acl-short.43/",
    doi = "10.18653/v1/2022.acl-short.43",
    pages = "385--393"
}

@inproceedings{
morphtok2024,
title={MorphTok: Morphologically Grounded Tokenization for Indic languages},
author={Maharaj Brahma and N J Karthika and Atul Kumar Singh and Devaraja Adiga and Smruti Bhate and Ganesh Ramakrishnan and Rohit Saluja and Maunendra Sankar Desarkar},
booktitle={Tokenization Workshop},
year={2025},
url={https://openreview.net/forum?id=32wHtvZOqH}
}

@INCOLLECTION{Lian2025-ch,
  title     = "Lbpe: Long-token-first tokenization to improve large language
               models",
  booktitle = "{ICASSP} 2025-2025 {IEEE} International Conference on Acoustics,
               Speech and Signal Processing ({ICASSP})",
  author    = "Lian, Haoran",
  publisher = "IEEE",
  year      =  2025
}

@inproceedings{seker-etal-2022-alephbert,
    title = "{A}leph{BERT}: Language Model Pre-training and Evaluation from Sub-Word to Sentence Level",
    author = "Seker, Amit  and
      Bandel, Elron  and
      Bareket, Dan  and
      Brusilovsky, Idan  and
      Greenfeld, Refael  and
      Tsarfaty, Reut",
    editor = "Muresan, Smaranda  and
      Nakov, Preslav  and
      Villavicencio, Aline",
    booktitle = "Proceedings of the 60th Annual Meeting of the Association for Computational Linguistics (Volume 1: Long Papers)",
    month = may,
    year = "2022",
    address = "Dublin, Ireland",
    publisher = "Association for Computational Linguistics",
    url = "https://aclanthology.org/2022.acl-long.4/",
    doi = "10.18653/v1/2022.acl-long.4",
    pages = "46--56"
}

@article{Toraman_2023,
   title={Impact of Tokenization on Language Models: An Analysis for Turkish},
   volume={22},
   ISSN={2375-4702},
   url={http://dx.doi.org/10.1145/3578707},
   DOI={10.1145/3578707},
   number={4},
   journal={ACM Transactions on Asian and Low-Resource Language Information Processing},
   publisher={Association for Computing Machinery (ACM)},
   author={Toraman, Cagri and Yilmaz, Eyup Halit and Sahinuc, Furkan and Ozcelik, Oguzhan},
   year={2023},
   month=mar, pages={1–21} }

@misc{haga2025babylmchallengeexploringeffect,
      title={BabyLM Challenge: Exploring the Effect of Variation Sets on Language Model Training Efficiency}, 
      author={Akari Haga and Akiyo Fukatsu and Miyu Oba and Arianna Bisazza and Yohei Oseki},
      year={2025},
      eprint={2411.09587},
      archivePrefix={arXiv},
      primaryClass={cs.CL},
      url={https://arxiv.org/abs/2411.09587}, 
}

@inproceedings{pennington2014glove,
  title={Glove: Global vectors for word representation},
  author={Pennington, Jeffrey and Socher, Richard and Manning, Christopher D},
  booktitle={Proceedings of the 2014 conference on empirical methods in natural language processing (EMNLP)},
  pages={1532--1543},
  year={2014}
}

@article{mikolov2013distributed,
  title={Distributed representations of words and phrases and their compositionality},
  author={Mikolov, Tomas and Sutskever, Ilya and Chen, Kai and Corrado, Greg S and Dean, Jeff},
  journal={Advances in neural information processing systems},
  volume={26},
  year={2013}
}

@article{sarzynska2021detecting,
  title={Detecting formal thought disorder by deep contextualized word representations},
  author={Sarzynska-Wawer, Justyna and Wawer, Aleksander and Pawlak, Aleksandra and Szymanowska, Julia and Stefaniak, Izabela and Jarkiewicz, Michal and Okruszek, Lukasz},
  journal={Psychiatry research},
  volume={304},
  pages={114135},
  year={2021},
  publisher={Elsevier}
}

@article{clark2022canine,
  title={Canine: Pre-training an efficient tokenization-free encoder for language representation},
  author={Clark, Jonathan H and Garrette, Dan and Turc, Iulia and Wieting, John},
  journal={Transactions of the Association for Computational Linguistics},
  volume={10},
  pages={73--91},
  year={2022},
  publisher={MIT Press One Rogers Street, Cambridge, MA 02142-1209, USA journals-info~…}
}

@article{deiseroth2024t,
  title={T-FREE: Subword tokenizer-free generative LLMs via sparse representations for memory-efficient embeddings},
  author={Deiseroth, Bj{\"o}rn and Brack, Manuel and Schramowski, Patrick and Kersting, Kristian and Weinbach, Samuel},
  journal={arXiv preprint arXiv:2406.19223},
  year={2024}
}

@article{team2025gemma,
  title={Gemma 3 technical report},
  author={Team, Gemma and Kamath, Aishwarya and Ferret, Johan and Pathak, Shreya and Vieillard, Nino and Merhej, Ramona and Perrin, Sarah and Matejovicova, Tatiana and Ram{\'e}, Alexandre and Rivi{\`e}re, Morgane and others},
  journal={arXiv preprint arXiv:2503.19786},
  year={2025}
}

@article{yang2025qwen3,
  title={Qwen3 technical report},
  author={Yang, An and Li, Anfeng and Yang, Baosong and Zhang, Beichen and Hui, Binyuan and Zheng, Bo and Yu, Bowen and Gao, Chang and Huang, Chengen and Lv, Chenxu and others},
  journal={arXiv preprint arXiv:2505.09388},
  year={2025}
}

@article{guo2025deepseek,
  title={Deepseek-r1: Incentivizing reasoning capability in llms via reinforcement learning},
  author={Guo, Daya and Yang, Dejian and Zhang, Haowei and Song, Junxiao and Zhang, Ruoyu and Xu, Runxin and Zhu, Qihao and Ma, Shirong and Wang, Peiyi and Bi, Xiao and others},
  journal={arXiv preprint arXiv:2501.12948},
  year={2025}
}

@article{brown2020language,
  title={Language models are few-shot learners},
  author={Brown, Tom and Mann, Benjamin and Ryder, Nick and Subbiah, Melanie and Kaplan, Jared D and Dhariwal, Prafulla and Neelakantan, Arvind and Shyam, Pranav and Sastry, Girish and Askell, Amanda and others},
  journal={Advances in neural information processing systems},
  volume={33},
  pages={1877--1901},
  year={2020}
}

@inproceedings{chelombitko2025samponlp,
  title={SampoNLP: A Self-Referential Toolkit for Morphological Analysis of Subword Tokenizers},
  author={Chelombitko, Iaroslav and Chelombitko, Ekaterina and Komissarov, Aleksey},
  booktitle={Proceedings of the 10th International Workshop on Computational Linguistics for Uralic Languages},
  pages={57--67},
  year={2025}
}

@article{asgari2025morphbpe,
  title={MorphBPE: A Morpho-Aware Tokenizer Bridging Linguistic Complexity for Efficient LLM Training Across Morphologies},
  author={Asgari, Ehsaneddin and Kheir, Yassine El and Javaheri, Mohammad Ali Sadraei},
  journal={arXiv preprint arXiv:2502.00894},
  year={2025}
}

@article{arnett2025evaluating,
  title={Evaluating morphological alignment of tokenizers in 70 languages},
  author={Arnett, Catherine and Hudspeth, Marisa and O'Connor, Brendan},
  journal={arXiv preprint arXiv:2507.06378},
  year={2025}
}

@article{sandhan2022translist,
  title={TransLIST: A transformer-based linguistically informed Sanskrit tokenizer},
  author={Sandhan, Jivnesh and Singha, Rathin and Rao, Narein and Samanta, Suvendu and Behera, Laxmidhar and Goyal, Pawan},
  journal={arXiv preprint arXiv:2210.11753},
  year={2022}
}

@article{gaikwad2021state,
  title={On state-of-the-art of POS tagger,‘Sandhi’Splitter,‘Alankaar’Finder and ‘Samaas’ Finder for Indo-Aryan and Dravidian languages},
  author={Gaikwad, Hema and Saini, Jatinderkumar R},
  journal={Int J Adv Comput Sci Appl},
  volume={12},
  number={4},
  pages={429--436},
  year={2021}
}

@article{xue2022byt5,
  title={ByT5: Towards a token-free future with pre-trained byte-to-byte models},
  author={Xue, Linting and Barua, Aditya and Constant, Noah and Al-Rfou, Rami and Narang, Sharan and Kale, Mihir and Roberts, Adam and Raffel, Colin},
  journal={Transactions of the Association for Computational Linguistics},
  volume={10},
  pages={291--306},
  year={2022},
  publisher={MIT Press One Broadway, 12th Floor, Cambridge, Massachusetts 02142, USA~…}
}

@article{tay2021charformer,
  title={Charformer: Fast character transformers via gradient-based subword tokenization},
  author={Tay, Yi and Tran, Vinh Q and Ruder, Sebastian and Gupta, Jai and Chung, Hyung Won and Bahri, Dara and Qin, Zhen and Baumgartner, Simon and Yu, Cong and Metzler, Donald},
  journal={arXiv preprint arXiv:2106.12672},
  year={2021}
}

@inproceedings{ali2024tokenizer,
  title={Tokenizer choice for llm training: Negligible or crucial?},
  author={Ali, Mehdi and Fromm, Michael and Thellmann, Klaudia and Rutmann, Richard and L{\"u}bbering, Max and Leveling, Johannes and Klug, Katrin and Ebert, Jan and Doll, Niclas and Buschhoff, Jasper and others},
  booktitle={Findings of the Association for Computational Linguistics: NAACL 2024},
  pages={3907--3924},
  year={2024}
}

@article{kumar2026sanskrit,
  title={Is Sanskrit the most token-efficient language? A quantitative study using GPT, Gemini, and SentencePiece},
  author={Kumar, Anshul},
  journal={arXiv preprint arXiv:2601.06142},
  year={2026}
}

@inproceedings{pattnayak2025tokenization,
  title={Tokenization matters: Improving zero-shot ner for indic languages},
  author={Pattnayak, Priyaranjan and Patel, Hitesh and Agarwal, Amit},
  booktitle={2025 IEEE International Conference on Electro Information Technology (eIT)},
  pages={456--462},
  year={2025},
  organization={IEEE}
}

@article{tamang2024evaluating,
  title={Evaluating Tokenizer Performance of Large Language Models Across Official Indian Languages},
  author={Tamang, Sagar and Bora, Dibya Jyoti},
  journal={arXiv preprint arXiv:2411.12240},
  year={2024}
}

@inproceedings{chaudhari2023significance,
  title={On Significance of Subword Tokenization for Low-Resource and Efficient Named Entity Recognition: A Case Study in Marathi},
  author={Chaudhari, Harsh and Patil, Anuja and Lavekar, Dhanashree and Khairnar, Pranav and Joshi, Raviraj and Pande, Sachin},
  booktitle={International Conference on Data Analytics \& Management},
  pages={483--494},
  year={2023},
  organization={Springer}
}

@misc{jabbar2024morphpiecelinguistictokenizer,
      title={MorphPiece : A Linguistic Tokenizer for Large Language Models}, 
      author={Haris Jabbar},
      year={2024},
      eprint={2307.07262},
      archivePrefix={arXiv},
      primaryClass={cs.CL},
      url={https://arxiv.org/abs/2307.07262}, 
}

@misc{batsuren2022unimorph40universalmorphology,
      title={UniMorph 4.0: Universal Morphology}, 
      author={Khuyagbaatar Batsuren et al.},
      year={2022},
      eprint={2205.03608},
      archivePrefix={arXiv},
      primaryClass={cs.CL},
      url={https://arxiv.org/abs/2205.03608}, 
}

@inproceedings{baxi-bhatt-2022-gujmorph,
    title = "{G}uj{MORPH} - A Dataset for Creating {G}ujarati Morphological Analyzer",
    author = "Baxi, Jatayu  and
      Bhatt, Brijesh",
    editor = "Calzolari, Nicoletta  and
      B{\'e}chet, Fr{\'e}d{\'e}ric  and
      Blache, Philippe  and
      Choukri, Khalid  and
      Cieri, Christopher  and
      Declerck, Thierry  and
      Goggi, Sara  and
      Isahara, Hitoshi  and
      Maegaard, Bente  and
      Mariani, Joseph  and
      Mazo, H{\'e}l{\`e}ne  and
      Odijk, Jan  and
      Piperidis, Stelios",
    booktitle = "Proceedings of the Thirteenth Language Resources and Evaluation Conference",
    month = jun,
    year = "2022",
    address = "Marseille, France",
    publisher = "European Language Resources Association",
    url = "https://aclanthology.org/2022.lrec-1.767/",
    pages = "7088--7095"
}

@inproceedings{batsuren-etal-2021-morphynet,
    title = "{M}orphy{N}et: a Large Multilingual Database of Derivational and Inflectional Morphology",
    author = "Batsuren, Khuyagbaatar  and
      Bella, G{\'a}bor  and
      Giunchiglia, Fausto",
    editor = "Nicolai, Garrett  and
      Gorman, Kyle  and
      Cotterell, Ryan",
    booktitle = "Proceedings of the 18th SIGMORPHON Workshop on Computational Research in Phonetics, Phonology, and Morphology",
    month = aug,
    year = "2021",
    address = "Online",
    publisher = "Association for Computational Linguistics",
    url = "https://aclanthology.org/2021.sigmorphon-1.5/",
    doi = "10.18653/v1/2021.sigmorphon-1.5",
    pages = "39--48"
}

@inproceedings{papineni-etal-2002-bleu,
    title = "{B}leu: a Method for Automatic Evaluation of Machine Translation",
    author = "Papineni, Kishore  and
      Roukos, Salim  and
      Ward, Todd  and
      Zhu, Wei-Jing",
    editor = "Isabelle, Pierre  and
      Charniak, Eugene  and
      Lin, Dekang",
    booktitle = "Proceedings of the 40th Annual Meeting of the Association for Computational Linguistics",
    month = jul,
    year = "2002",
    address = "Philadelphia, Pennsylvania, USA",
    publisher = "Association for Computational Linguistics",
    url = "https://aclanthology.org/P02-1040/",
    doi = "10.3115/1073083.1073135",
    pages = "311--318"
}

@inproceedings{popovic-2015-chrf,
    title = "chr{F}: character n-gram {F}-score for automatic {MT} evaluation",
    author = "Popovi{\'c}, Maja",
    editor = "Bojar, Ond{\v{r}}ej  and
      Chatterjee, Rajan  and
      Federmann, Christian  and
      Haddow, Barry  and
      Hokamp, Chris  and
      Huck, Matthias  and
      Logacheva, Varvara  and
      Pecina, Pavel",
    booktitle = "Proceedings of the Tenth Workshop on Statistical Machine Translation",
    month = sep,
    year = "2015",
    address = "Lisbon, Portugal",
    publisher = "Association for Computational Linguistics",
    url = "https://aclanthology.org/W15-3049/",
    doi = "10.18653/v1/W15-3049",
    pages = "392--395"
}

@inproceedings{ott2019fairseq,
  title = {fairseq: A Fast, Extensible Toolkit for Sequence Modeling},
  author = {Myle Ott and Sergey Edunov and Alexei Baevski and Angela Fan and Sam Gross and Nathan Ng and David Grangier and Michael Auli},
  booktitle = {Proceedings of NAACL-HLT 2019: Demonstrations},
  year = {2019},
}

@article{jain2024bhasaanuvaad,
  title   = {BhasaAnuvaad: A Speech Translation Dataset for 14 Indian Languages},
  author  = {Sparsh Jain and Ashwin Sankar and Devilal Choudhary and Dhairya Suman and Nikhil Narasimhan and Mohammed Safi Ur Rahman Khan and Anoop Kunchukuttan and Mitesh M Khapra and Raj Dabre},
  year    = {2024},
  journal = {arXiv preprint arXiv: 2411.04699}
}

@misc{liu2025superbpespacetravellanguage,
      title={SuperBPE: Space Travel for Language Models}, 
      author={Alisa Liu and Jonathan Hayase and Valentin Hofmann and Sewoong Oh and Noah A. Smith and Yejin Choi},
      year={2025},
      eprint={2503.13423},
      archivePrefix={arXiv},
      primaryClass={cs.CL},
      url={https://arxiv.org/abs/2503.13423}, 
}

@article{isac2025slip,
  title={SLIP: A Sanskrit Linguistic Intelligence Pipeline for Enhanced Neural Machine Translation of Classical Texts},
  author={Isac, N Biraja and Das, Himansu},
  journal={IEEE Access},
  year={2025},
  publisher={IEEE}
}

@misc{mikolov2013efficientestimationwordrepresentations,
      title={Efficient Estimation of Word Representations in Vector Space}, 
      author={Tomas Mikolov and Kai Chen and Greg Corrado and Jeffrey Dean},
      year={2013},
      eprint={1301.3781},
      archivePrefix={arXiv},
      primaryClass={cs.CL},
      url={https://arxiv.org/abs/1301.3781}, 
}

@misc{charlet_2025_agbpe_v3,
  author       = {Charlet, Théo},
  title        = {AG-BPE: Attention-Guided Byte-Pair Encoding 
                  for Semantic-Aware Tokenization},
  month        = jul,
  year         = 2025,
  doi          = {10.5281/zenodo.15864340},
  url          = {https://doi.org/10.5281/zenodo.15864340}
}

\clearpage
\newpage

\onecolumn
\setcounter{page}{1}
\centerline{\textbf{\LARGE \textsc{SuTRA}: Supplementary Material}}


\vspace{2em}
\hrule
\vspace{1em}
\noindent \textbf{\large Summary of Supplementary Contents}
\vspace{0.5em}
\begin{enumerate}[label=Section \arabic*:, leftmargin=5em]
    \item \textbf{Methodology for Morphological Dataset Preparation} (Page \pageref{sec:data_method}) \\ 
    Details on the three-stage pipeline (SampoNLP, statistical filtering, and LLM-based verification) used to construct the Gold Standard dataset.
    
    \item \textbf{The Rigidity Constraint ($\gamma$)} (Page \pageref{sec:gamma_ablation}) \\ 
    Formal derivation of the dynamic annealing schedule and the linguistic justification for the curriculum-based merging approach.
    
    \item \textbf{Algorithm for \textsc{SuTRA}} (Page \pageref{alg:sutra_final}) \\ 
    Complete pseudocode for the Semantic-Unit Tokenization with Rigidity Annealing process, including pre-tokenization and trilateral enforcement.
    
    \item \textbf{Computational Complexity Analysis} (Page \pageref{sec:complexity}) \\ 
    A theoretical comparison of training and inference time complexities between \textsc{SuTRA}, standard \textsc{Bpe}, and \textsc{MorphTok}.
    
    \item \textbf{Extended Analysis of Morphological Alignment} (Page \pageref{tab:tok_results}) \\ 
    Full evaluation metrics (Exact Match, Precision, Recall) across all target languages, expanding on the main paper's alignment results.
    
    \item \textbf{Experimental Setup for Neural Machine Translation} (Page \pageref{sec:mt_setup_supp}) \\ 
    Detailed architectural configurations, optimization hyperparameters, and data preprocessing protocols for the translation tasks.
    
    \item \textbf{Experiment: Causal Language Modeling} (Page \pageref{tab:clm_results}) \\ 
    Evaluation of predictive efficiency, perplexity derivations, and training loss curves for generative modeling.
    
    \item \textbf{Extended Analysis of Morphological Robustness} (Page \pageref{tab:baseline_comparison}) \\ 
    A deep dive into the "Morphological Shattering" phenomenon and secondary metrics quantifying tokenizer resilience under noise.
    
    \item \textbf{Limitations} (Page \pageref{sec:limitations}) \\ 
    Discussion on vocabulary scaling and the boundaries of semantic constraint saturation.
\end{enumerate}
\vspace{1em}
\hrule
\vspace{2em}

\noindent\section{Methodology for Morphological Dataset Preparation}
\label{sec:data_method}

We developed a robust, three-stage pipeline to construct the Gold Standard Morphological Split Dataset. Our approach integrates unsupervised information-theoretic segmentation with semantic verification using \textbf{Gemini 3.0 Flash}~\cite{team2023gemini}. This hybrid methodology allows us to leverage the high recall of statistical segmentation while ensuring precision through expert-aligned semantic reasoning.

\subsection{Stage 1: Adapted Iterative Morphological Decomposition (SampoNLP)}
\label{subsec:samponlp}

To generate initial candidate splits, we adapted the \textbf{SampoNLP} pipeline \cite{chelombitko2025samponlp}. This framework operates on the principle of \textit{Self-Referential Atomicity Scoring}, where the "atomic" nature of a morpheme is determined iteratively by its ability to optimally compress the lexicon.

While the core logic is language-agnostic, the original implementation relied on Latin-script constraints. We adapted the pipeline for Indic languages (\textit{Hindi}, \textit{Marathi}, \textit{Gujarati}) through the following modifications:

\begin{enumerate}
    \item \textbf{Script-Specific Hard Filtering:} We redefined the valid character set $\Sigma$ to include the Unicode blocks for Devanagari (\texttt{U+0900}--\texttt{U+097F}) and Gujarati (\texttt{U+0A80}--\texttt{U+0AFF}).
    \item \textbf{Atomic Whitelist Expansion ($W$):} We significantly expanded the whitelist $W$ to include \textbf{Matras (Dependent Vowels)} (e.g., \textit{\textipa{A}} (\texttt{U+093E})) and short grammatical suffixes. This ensures that legitimate single-character morphemes are not aggressively merged into adjacent roots.
\end{enumerate}

\subsection{Stage 2: Limitations of Statistical Segmentation}
\label{subsec:stat_limits}

While the adapted pipeline successfully identified high-frequency recurring patterns, we observed a systematic error pattern defined as \textbf{Statistical Over-segmentation}. The algorithm would frequently split a root word if it contained a substring resembling a high-frequency suffix (e.g., splitting Hindi \textit{shriman} (Mr.) into \textit{shri}+\textit{man}), ignoring the semantic integrity of the root. This necessitated a semantic verification layer.

\subsection{Stage 3: LLM-Based Semantic Verification}
\label{subsec:llm_verification}

To resolve over-segmentation and restore oblique roots, we employed \textbf{Gemini 3.0 Flash}~\cite{team2023gemini}. We utilized a unified system prompt template across all languages, dynamically injecting language-specific morphological rules to handle the distinction between agglutinative Indic scripts and fusional English morphology.

\begin{tcolorbox}[colback=gray!5,colframe=gray!40,title=Unified System Prompt Template Used for Verification]
\small
\textbf{Role:} You are an expert linguist specializing in \texttt{[Target\_Language]} Morphology.\\
\textbf{Task:} Your task is to segment the given word into its constituent morphemes: the Root and any Suffixes.

\texttt{[Target\_Language]} is a morphologically rich language. Pay close attention to:
\texttt{[Language\_Specific\_Rules]}\\

Return the answer in strict JSON format:
\begin{verbatim}
{
  "is_correct": boolean,
  "correct_split": "Prefix+Root+Suffix", 
  "morphology": {
    "root": "The base dictionary form (e.g., 'ghar', 'happy')",
    "prefixes": [],
    "suffixes": ["grammatical suffixes"],
    "stem": "Modified stem if applicable"
  },
  "origin": "Native [Target_Language]",
  "error_type": "None"
}
\end{verbatim}
\end{tcolorbox}

\begin{tcolorbox}[colback=white,colframe=gray!60,title=Language-Specific Rule Injections]
\small

\textbf{Marathi / Hindi Rules:}
\begin{enumerate}
    \item \textbf{Oblique forms (Samanya Rup):} Restore the root to its original dictionary form (e.g., 'ghara' $\rightarrow$ Root: 'ghar', Suffix: 'cha').
    \item \textbf{Plural markers.}
    \item \textbf{Case markers (Vibhakti).}
\end{enumerate}

\par\medskip\hrule\medskip\par

\textbf{Gujarati Rules:}
\begin{enumerate}
    \item \textbf{Root Identification:} Identify the base dictionary form (e.g., for 'gharnu', the root is 'ghar').
    \item \textbf{Oblique Forms:} Restore the root if it changed form before a suffix was added (e.g., 'chokraao' $\rightarrow$ Root: 'chokro', Suffix: 'ao').
    \item \textbf{Suffixes:} Isolate grammatical markers for case (nu/ni/no/na), plurality (o), and postpositions.
\end{enumerate}

\par\medskip\hrule\medskip\par

\textbf{English Rules:}
\begin{enumerate}
    \item \textbf{Root Restoration:} Restore the root's original spelling if it changed during affixation.
    \begin{itemize}
        \item Example: 'happiness' $\rightarrow$ Root: 'happy' (NOT 'happi').
        \item Example: 'running' $\rightarrow$ Root: 'run' (NOT 'runn').
    \end{itemize}
    \item \textbf{Affixation:} Identify standard prefixes (e.g., 'un-', 're-') and suffixes (e.g., '-ing', '-ed').
    \item \textbf{Compound Words:} If the word is a compound, split it into its constituent free bases.
\end{enumerate}

\end{tcolorbox}

\textbf{Justification for the Prompt Design:}
The prompt explicitly enforces \textbf{Root Restoration}. Standard statistical segmenters operate strictly on surface forms and cannot recover the original dictionary root (e.g., mapping *happi* back to *happy* or *ghara* back to *ghar*). By enforcing a structured JSON output with a separate field for the "root", we ensure that the resulting dataset captures the deep morphological structure, not just surface segmentation.

\newpage

\section{The Rigidity Constraint ($\gamma$)}
\label{sec:gamma_ablation}

In the \textsc{SuTRA} framework, the merging process is governed by a dynamic Rigidity Constraint ($\gamma_t$), which acts as an exponential penalty on the semantic validity probability $\Psi(a,b)$ of a candidate merge pair. The score is calculated as $S(a,b) = f(a,b) \cdot \Psi(a,b)^{\gamma_t}$. 

\noindent\textbf{Annealing Formulation.} To transition the vocabulary construction from strict morphological adherence to optimal statistical compression, we anneal $\gamma_t$ linearly over the course of the \textsc{Bpe} training steps. Let $t$ represent the current merge iteration and $T$ represent the total number of target merges to reach the desired vocabulary size. The constraint at step $t$ is defined as:
$$ \gamma_t = \gamma_{start} - \left( \gamma_{start} - \gamma_{end} \right) \cdot \left( \frac{t}{T} \right) $$

\noindent\textbf{Justification for the Curriculum Approach.} The intuition behind this dynamic decay is rooted in curriculum learning. Early in the \textsc{Bpe} training process, the algorithm combines short character sequences to form the foundational roots and affixes of the language. If these foundational units are permitted to cross morphological boundaries, the error propagates, causing the \textit{Morphological Shattering} observed in standard tokenizers. By enforcing a high rigidity constraint early on, we force the algorithm to build "safe," semantically pure subwords. As the vocabulary grows ($t \to T$), the algorithm needs to form longer, composite tokens (e.g., highly frequent full words) to ensure statistical efficiency and maintain a competitive compression ratio. Relaxing the constraint allows these necessary larger merges to occur once the foundational roots are already securely established.

\noindent\textbf{Why $\gamma \in [4, 0]$?} We specifically tune the annealing schedule to bound $\gamma_t$ between $4$ and $0$.
\begin{itemize}
    \item \textbf{Starting at $\gamma_{start} = 4$:} Because the semantic validity $\Psi(a,b)$ is a probability between $0$ and $1$, raising it to the power of $4$ acts as an aggressive, non-linear amplifier for any boundary conflicts. For example, if a candidate merge crosses a forbidden boundary in just $10\%$ of its occurrences ($\Psi = 0.9$), its score is severely penalized ($\Psi^4 \approx 0.65$), effectively blocking the merge in favor of structurally safer alternatives.
    \item \textbf{Ending at $\gamma_{end} = 0$:} As $\gamma_t$ approaches $0$, the penalty factor $\Psi(a,b)^0$ approaches $1$. At the very last merge step ($t = T$), the scoring function cleanly reduces to $S(a,b) = f(a,b) \cdot 1$, which is the exact formulation of standard \textsc{Bpe}. This ensures that the final stages of vocabulary construction are purely frequency-driven, allowing \textsc{SuTRA} to match the fertility and sequence compression efficiency of baseline tokenizers without sacrificing the structural integrity built during the early phases.
\end{itemize}
\begin{table}[ht!]
\centering
\small
\resizebox{\columnwidth}{!}{%
\begin{tabular}{l | cc | c c}
\toprule
\multirow{2}{*}{\textbf{Candidate Merge $(a, b)$}} & \multirow{2}{*}{\textbf{Freq.} $f$} & \multirow{2}{*}{\textbf{Validity} $\Psi$} & \multicolumn{2}{c}{\textbf{Effective Score} $S(a,b)$} \\
\cmidrule(lr){4-5}
& & & \textbf{$t=0$ ($\gamma=4$)} & \textbf{$t=T$ ($\gamma=0$)} \\
\midrule
\textit{Safe Merge} (e.g., intra-root) & $1000$ & $1.00$ & $\mathbf{1000}$ & $1000$ \\
\textit{Boundary Violation} (e.g., prefix+root) & $1200$ & $0.80$ & $491$ & $\mathbf{1200}$ \\
\textit{Severe Violation} (e.g., shattered prefix) & $1500$ & $0.50$ & $93$ & $\mathbf{1500}$ \\
\bottomrule
\end{tabular}%
}
\caption{\textbf{Effect of the Annealed Rigidity Constraint.} At the start of training ($t=0, \gamma=4$), the exponential penalty prioritizes structurally safe merges ($\Psi \approx 1.0$), forcing the tokenizer to build valid semantic roots despite lower raw frequencies. By the end of training ($t=T, \gamma=0$), the penalty disappears ($\Psi^0 = 1$), reducing the score to standard \textsc{Bpe} frequency to optimize compression.}
\label{tab:gamma_example}
\end{table}

\newpage

\section{Algorithm for \textsc{SuTRA}}

\begin{algorithm}[ht]
\caption{SuTRA: Semantic-Unit Tokenization with Rigidity Annealing}\label{alg:sutra_final}
\begin{algorithmic}[1]
\Require Corpus $\mathcal{C}$, Target Vocabulary Size $K$, Annealing Bounds $[\gamma_{start}, \gamma_{end}]$
\Ensure Final Vocabulary $\mathcal{V}$, Merge Rules $\mathcal{R}$

\Statex \textbf{Phase 1: Pre-tokenization \& Boundary Identification}
\State $\mathcal{C}' \gets \text{ApplyOrthographicRules}(\mathcal{C}, \Phi)$ \Comment{Group chars into Aksharas}
\State $\mathcal{V}_0 \gets \text{UniqueUnits}(\mathcal{C}')$
\State $\mathcal{B}_{forbidden} \gets \text{IdentifyMorphologicalBoundaries}(\mathcal{C})$ \Comment{Lexicon/Seq2Seq}

\Statex \textbf{Phase 2: Morphology-Aware Merging (Curriculum Learning)}
\State $\mathcal{V} \gets \mathcal{V}_0, \mathcal{R} \gets \emptyset$
\While{$|\mathcal{V}| < K$}
    \State $p \gets \frac{|\mathcal{V}| - |\mathcal{V}_0|}{K - |\mathcal{V}_0|}$ \Comment{Training Progress}
    \State $\gamma_t \gets \gamma_{start} \cdot (\frac{\gamma_{end}}{\gamma_{start}})^p$ \Comment{Exponential Annealing}
    
    \State \textbf{Step A: Collect Statistics}
    \ForAll{adjacent pairs $(a,b) \in \mathcal{C}'$}
        \State $f(a,b) \gets \text{CountFreq}(a,b)$
        \State $\chi(a,b) \gets \text{CountForbiddenBoundaryCrossings}(a,b, \mathcal{B}_{forbidden})$
    \EndFor
    
    \State \textbf{Step B: Score and Merge}
    \State $\Psi(a,b) \gets 1 - \frac{\chi(a,b)}{f(a,b)}$ \Comment{Morphological Validity}
    \State $(a^*, b^*) \gets \arg\max \left( f(a,b) \cdot \Psi(a,b)^{\gamma_t} \right)$
    
    \State $\mathcal{V} \gets \mathcal{V} \cup \{a^*b^*\}$
    \State $\mathcal{R} \gets \mathcal{R} \cup \{(a^*, b^*) \to a^*b^*\}$
    \State $\mathcal{C}' \gets \text{ApplyMerge}(\mathcal{C}', a^*, b^*)$
\EndWhile

\Statex \textbf{Phase 3: Structural Bottleneck (Post-processing)}
\ForAll{word $w \in \mathcal{C}'$}
    \If{$\text{SegmentCount}(w) > 3$} 
        \State $\text{IdentifyRootIndices}(w, \mathcal{B}_{forbidden})$
        \State $w \gets \text{ForceMergeRootTokens}(w)$ \Comment{Enforce $W = [P] \oplus R \oplus [S]$}
    \EndIf
\EndFor
\end{algorithmic}
\end{algorithm}



\section{Computational Complexity Analysis for \textsc{SuTRA}}
\label{sec:complexity}
This section details the time complexity of \textsc{SuTRA} during vocabulary construction (training) and text encoding (inference), compared against standard \textsc{Bpe} and \textsc{MorphTok}. Let $N$ be the total characters in the training corpus, $V$ be the target vocabulary size (number of merges), $M$ be the number of unique active candidate pairs, $V_{in}$ be the number of unique words in the training corpus, and $|w|$ be the length of a given word.

We utilize ByT5 for morphological boundary prediction in the baseline \textsc{MorphTok} and the pre-segmentation phase of \textsc{SuTRA}. As a byte-level Transformer, ByT5 exhibits quadratic time complexity with respect to sequence length, yielding an inference complexity of $\mathcal{O}(|w|^2)$ per word.

\subsection{Training Complexity}
\noindent\textbf{Standard \textsc{Bpe}:} The initial character pair frequency calculation takes $\mathcal{O}(N)$. Executing $V$ merges requires updating the frequencies of adjacent pairs, taking $\mathcal{O}(\log M)$ per update. The total training complexity is $\mathcal{O}(N + V \log M)$.

\noindent\textbf{\textsc{MorphTok}:} \textsc{MorphTok} requires pre-segmenting the corpus. Generating boundaries for $V_{in}$ unique words using ByT5 takes $\mathcal{O}(V_{in} \cdot |w|^2)$. After segmentation, standard \textsc{Bpe} is executed within the constrained boundaries. The total training complexity is $\mathcal{O}(V_{in} \cdot |w|^2 + N + V \log M)$.

\noindent\textbf{\textsc{SuTRA} (Ours):} \textsc{SuTRA} involves two primary training phases:
\begin{enumerate}
    \item \textbf{Boundary Extraction:} Similar to \textsc{MorphTok}, extracting boundaries for unique corpus words via ByT5 takes $\mathcal{O}(V_{in} \cdot |w|^2)$.
    \item \textbf{Score Calculation and Merging:} Initializing the boundary conflict counts $\chi(a,b)$ and valid frequencies $f(a,b)$ across the corpus takes $\mathcal{O}(N)$. During each of the $V$ merges, the algorithm updates the semantic validity probability $\Psi(a,b)$ and calculates the annealed score $S(a,b) = f(a,b) \cdot \Psi(a,b)^{\gamma_t}$. The exponentiation and scalar multiplication add an $\mathcal{O}(1)$ arithmetic operation per pair update. The merge phase therefore remains bounded by $\mathcal{O}(V \log M)$.
\end{enumerate}
The total training complexity for \textsc{SuTRA} is $\mathcal{O}(V_{in} \cdot |w|^2 + N + V \log M)$, which is asymptotically identical to \textsc{MorphTok}.

\subsection{Inference Complexity}
\noindent\textbf{Standard \textsc{Bpe}:} Encoding a word applies learned merge rules iteratively. The inference complexity is $\mathcal{O}(|w|)$.

\noindent\textbf{\textsc{MorphTok}:} \textsc{MorphTok} enforces boundaries at runtime. Unseen words must pass through the ByT5 model before subword merging can occur. The inference complexity is dominated by the Transformer forward pass, resulting in $\mathcal{O}(|w|^2)$ per out-of-vocabulary word.

\noindent\textbf{\textsc{SuTRA} (Ours):} \textsc{SuTRA} incorporates the morphological constraints directly into the final vocabulary merge rules during training. At inference, boundary prediction and score calculations are omitted. The model relies entirely on the learned merges, resulting in an inference complexity of $\mathcal{O}(|w|)$.

\begin{table}[htb]
\centering
\caption{\textbf{Time Complexity Summary.} Comparison of training and inference time complexity. Variables: $N$ (corpus size), $V$ (vocab size), $M$ (unique candidate pairs), $V_{in}$ (unique corpus words), and $|w|$ (word length).}
\label{tab:complexity}
\begin{tabular}{lcc}
\toprule
\textbf{Tokenizer} & \textbf{Training} & \textbf{Inference (per word)} \\
\midrule
Standard \textsc{Bpe} & $\mathcal{O}(N + V \log M)$ & $\mathcal{O}(|w|)$ \\
\textsc{MorphTok} & $\mathcal{O}(V_{in} \cdot |w|^2 + N + V \log M)$ & $\mathcal{O}(|w|^2)$ \\
\textbf{\textsc{SuTRA} (Ours)} & $\mathcal{O}(V_{in} \cdot |w|^2 + N + V \log M)$ & $\mathcal{O}(|w|)$ \\
\bottomrule
\end{tabular}
\end{table}

\newpage

\section{Extended Analysis of Morphological Alignment.} 

The streamlined results in Table \ref{tab:tok_results} highlight the tension between subword compression (Fertility) and linguistic integrity (Boundary F1):

\begin{table}[ht!]
\centering
\caption{Comparison of Tokenizers (Normalized) across Hindi, Marathi, and Gujarati. EM: Exact Match Acc, Prec: Boundary Precision, Rec: Boundary Recall, F1: Boundary F1, Fert: Fertility Ratio. Best F1 scores are in bold.}
\label{tab:tok_results}
\small
\setlength{\tabcolsep}{3pt}
\resizebox{\columnwidth}{!}{%
\begin{tabular}{l|ccccc|ccccc|ccccc}
\hline
\textbf{Tokenizer} & \multicolumn{5}{c|}{\textbf{Hindi}} & \multicolumn{5}{c|}{\textbf{Marathi}} & \multicolumn{5}{c|}{\textbf{Gujarati}} \\
\cline{2-16}
& \textbf{EM} & \textbf{Prec.} & \textbf{Rec.} & \textbf{F1} & \textbf{Fert.} & \textbf{EM} & \textbf{Prec.} & \textbf{Rec.} & \textbf{F1} & \textbf{Fert.} & \textbf{EM} & \textbf{Prec.} & \textbf{Rec.} & \textbf{F1} & \textbf{Fert.} \\
\hline
BPE       & 0.271 & 0.398 & 0.609 & 0.482 & 1.285 & 0.268 & 0.396 & 0.578 & 0.470 & 1.225 & 0.396 & 0.534 & 0.662 & 0.591 & 1.126 \\
SentencePiece & 0.191 & 0.357 & 0.566 & 0.438 & 1.315 & 0.000 & 0.043 & 1.000 & 0.083 & 1.183 & 0.378 & 0.523 & 0.679 & 0.591 & 1.156 \\
SuperBPE      & 0.000 & 0.056 & 0.215 & 0.089 & 2.517 & 0.000 & 0.050 & 0.264 & 0.084 & 3.084 & 0.000 & 0.058 & 0.289 & 0.096 & 3.113 \\
Unigram       & 0.185 & 0.359 & 0.566 & 0.439 & 1.310 & 0.394 & 0.511 & 0.778 & 0.507 & 1.256 & 0.460 & 0.600 & 0.756 & \textbf{0.669} & 1.137 \\
WordPiece     & 0.166 & 0.353 & 0.493 & 0.411 & 1.214 & 0.274 & 0.427 & 0.688 & 0.527 & 1.300 & 0.378 & 0.522 & 0.694 & 0.596 & 1.173 \\
MorphTok      & 0.004 & 0.122 & 0.531 & 0.190 & 2.601 & 0.008 & 0.126 & 0.763 & 0.216 & 3.482 & 0.006 & 0.145 & 0.835 & 0.247 & 3.494 \\
\textsc{SuTRA}         & 0.212 & 0.459 & 0.810 & \textbf{0.586} & 1.412 & 0.132 & 0.354 & 0.898 & \textbf{0.617} & 1.755 & 0.213 & 0.448 & 0.836 & 0.584 & 1.454 \\
\hline
\end{tabular}
}
\end{table}

\begin{enumerate}
    \item \textbf{Efficiency-Accuracy Frontier:} \textsc{SuTRA} (Ours) achieves the highest \textbf{Boundary F1} in Hindi ($0.586$) and Marathi ($0.617$), establishing a new performance baseline. Notably, it maintains high alignment scores despite a slightly higher fertility ratio than standard BPE. This suggests that the additional segments produced by \textsc{SuTRA} are not "noise" but rather accurate captures of functional affixes that purely statistical methods often miss.
    
    \item \textbf{The Morphological Shattering Threshold:} We observe a clear inverse correlation between extreme fertility and boundary alignment. Models that "shatter" words into excessive units, such as \textsc{MorphTok} and \textsc{SuperBPE} (Fert. $> 2.5$), consistently yield the lowest F1 scores ($< 0.25$). This confirms that over-segmentation in Indic scripts typically results in linguistically void fragments rather than meaningful sub-morphemic units.
    
    \item \textbf{Statistical Instability in Marathi:} The catastrophic failure of \textsc{SentencePiece} in Marathi (Fertility: $1.183$, F1: $0.083$) underscores the risk of purely data-driven tokenization. By reverting to character-level splitting, the model loses all semantic anchoring. In contrast, the phonetic guardrails in \textsc{SuTRA} ensure stable segmentation ($0.617$ F1) even in the presence of complex Marathi conjuncts.
    
    \item \textbf{Unigram Robustness in Gujarati:} In Gujarati, \textsc{Unigram} achieves a peak F1 of $0.669$ with the lowest fertility ($1.137$). This suggests that for certain scripts, the probabilistic pruning of a large initial vocabulary may capture stable roots more efficiently than the bottom-up merging approach of BPE-based architectures.
\end{enumerate}

These results validate that integrating orthographic constraints ($\Phi$) and morphological priors ($\Psi$) creates a more robust \textit{semantic anchor} for Indic LLMs, effectively mitigating the \textit{Morphological Shattering} prevalent in purely data-driven baselines.
\newpage 
\section{Experimental Setup for Neural Machine Translation}

\subsection{Neural Machine Translation}
\label{sec:mt_setup_supp}

We evaluate the downstream utility of \textsc{SuTRA} via a \textit{Hindi} $\rightarrow$ \textit{Marathi} translation task. Our framework is designed to isolate the impact of tokenization by keeping architectural and optimization variables constant across all baseline comparisons.

\subsubsection{Data and Preprocessing}
We utilize the \textit{BhasaAnuvaad} parallel corpus \cite{jain2024bhasaanuvaad} , employing official train (581379), validation (32298), and test (32300) splits. All tokenizer variants are trained and evaluated on identical sentence pairs to ensure a controlled environment.
\begin{itemize}
    \item \textbf{Normalization:} Before tokenization, we apply Unicode canonicalization and language-specific normalization via \textit{IndicNLP} to reduce orthographic variance.
    \item \textbf{Vocabularies:} We target a vocabulary size of 32k units per language. Data is binarized using \texttt{fairseq-preprocess} \cite{ott2019fairseq} with a shared subword dictionary between \textit{Hindi} and \textit{Marathi}.
\end{itemize}

\begin{figure}[ht]
    \centering
    \includegraphics[width=0.9\linewidth]{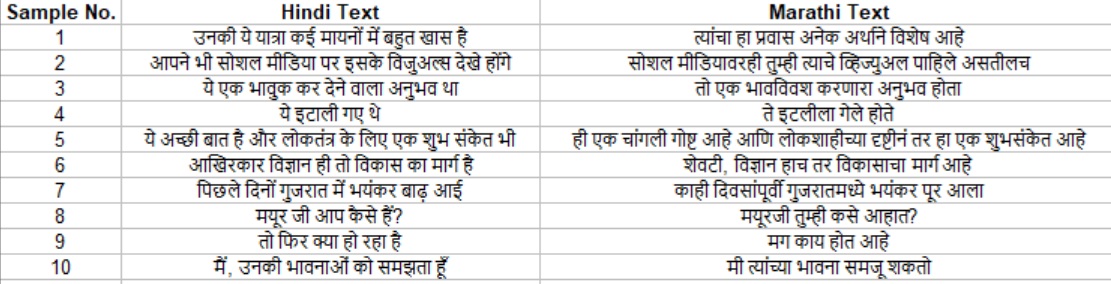}
    \caption{Sample examples from the dataset used for Hindi to Marathi Machine Translation Task}
    \label{fig:placeholder}
\end{figure}

\subsubsection{Model Architecture}
We employ a standard Transformer encoder--decoder with the following configuration:
\begin{itemize}
    \item \textbf{Depth and Width:} 3 encoder and 3 decoder layers; embedding dimension $d_{\mathrm{model}} = 100$; feed-forward dimension $d_{\mathrm{ff}} = 400$.
    \item \textbf{Attention:} 4 attention heads per layer with a dropout rate of 0.2.
    \item \textbf{Parameter Tying:} Decoder input and output embeddings are tied to improve convergence and reduce parameter count.
\end{itemize}

\subsubsection{Training and Optimization}
Models are trained for 100k updates using the Adam optimizer ($\beta_1=0.9, \beta_2=0.98$) with a label smoothing factor of 0.1. We utilize an inverse-square-root learning rate schedule with a peak of $5 \times 10^{-4}$ and 4,000 warmup updates. Gradient clipping is enforced at 1.0, and we use a token-based batch size of 4,096 per update. Best models are selected based on validation loss.

\subsubsection{Evaluation Protocol}
Final performance is measured on detokenized outputs using beam search (size 5) and a length penalty of 1.0. We report:
\begin{itemize}
    \item \textbf{Translation Quality:} \textsc{Bleu} \cite{papineni-etal-2002-bleu} and \textit{chrF} \cite{popovic-2015-chrf} scores computed on detokenized text.
    \item \textbf{Intrinsic Metrics:} Validation Perplexity (\textsc{Ppl}), token \textit{fertility} ratios, and Out-of-Vocabulary (\textsc{Oov}) percentages.
\end{itemize}

\newpage 

\section{Experiment Causal Language Modeling}

\subsection{Metric Derivation: Perplexity}
For the downstream Causal Language Modeling (CLM) task, we evaluate the predictive efficiency of \textsc{SuTRA} using \textbf{Perplexity (PPL)}. Perplexity measures the model's uncertainty when predicting the next token in a sequence. Formally, for a test sequence of tokens $W = (w_1, w_2, \dots, w_N)$, the perplexity is derived from the total \textbf{Negative Log-Likelihood (NLL)}.

Let $P(w_i \mid w_{<i})$ be the probability assigned by the model to the $i$-th token given the preceding context. The total NLL for the sequence is:
\begin{equation}
    \text{NLL}_{\text{total}} = - \sum_{i=1}^{N} \log P(w_i \mid w_{<i})
\end{equation}

In our implementation, we calculate the average NLL per predicted token to account for varying sequence lengths:
\begin{equation}
    \text{Avg NLL} = \frac{1}{N_{\text{pred}}} \text{NLL}_{\text{total}}
\end{equation}
where $N_{\text{pred}}$ is the count of non-padding tokens. The Perplexity is then defined as the exponential of the average NLL:
\begin{equation}
    PPL = \exp(\text{Avg NLL}) = e^{\left( \frac{1}{N_{\text{pred}}} \text{NLL}_{\text{total}} \right)}
\end{equation}

\subsection{Implementation of \textsc{SuTRA} Pre-tokenization}
To ensure morphological grounding for Indic scripts, our \texttt{SCBPEPreTokenizer} utilizes a regex-based grouping strategy before BPE merging. The pattern targets phonetic clusters (aksharas) across Devanagari and Gujarati Unicode blocks:
\begin{itemize}
    \item \textbf{Devanagari:} \texttt{\textbackslash u0900-\textbackslash u097F} (Base) and \texttt{\textbackslash u093A-\textbackslash u094F} (Marks).
    \item \textbf{Gujarati:} \texttt{\textbackslash u0A80-\textbackslash u0AFF} (Base) and \texttt{\textbackslash u0ABC-\textbackslash u0ACD} (Marks).
\end{itemize}
This ensures that dependent vowels (matras), halants, and nasalization marks (\textit{anusvara}) are never separated from their base consonant during the initial tokenization phase. Additionally, a word-boundary marker \texttt{</w>} is appended to the final phonetic unit of every word to preserve lexical boundaries.

\subsection{Model Configuration and Training Hyperparameters}
We trained a GPT-2 style Transformer architecture from scratch using the Hugging Face \texttt{Trainer} API. The detailed hyperparameters are provided in Table~\ref{tab:hyperparams}.

\begin{table}[h]
\centering
\small
\begin{tabular}{lp{4cm}}
\hline
\textbf{Hyperparameter} & \textbf{Value} \\ \hline
Architecture & GPT-2 (124M parameters) \\
Number of Layers & 12 \\
Number of Heads & 12 \\
Embedding Dimension ($d_{model}$) & 768 \\
Context Window ($n_{ctx}$) & 1024 \\
Vocabulary Size & Language-dependent \\ \hline
Optimizer & AdamW \\
Learning Rate & $5 \times 10^{-4}$ \\
Weight Decay & 0.01 \\
Warmup Steps & 1,000 \\
Target Training Tokens & 1.5 Billion \\
Batch Size (per device) & 16 \\
Precision & Mixed (\texttt{bf16} or \texttt{fp16}) \\ \hline
\end{tabular}
\caption{Hyperparameters for the downstream Causal Language Modeling task.}
\label{tab:hyperparams}
\end{table}

\subsection{Evaluation Protocol}
Out-of-distribution (OOD) perplexity was calculated on held-out parquet datasets. To ensure robustness, we used a row-batch size of 256 and a sequence length limit of 256 tokens. The model was evaluated in \texttt{eval()} mode with \texttt{torch.no\_grad()} to ensure deterministic NLL calculations.

\subsection{Results}
\begin{figure}[ht]
    \centering
    \includegraphics[width=0.95\linewidth]{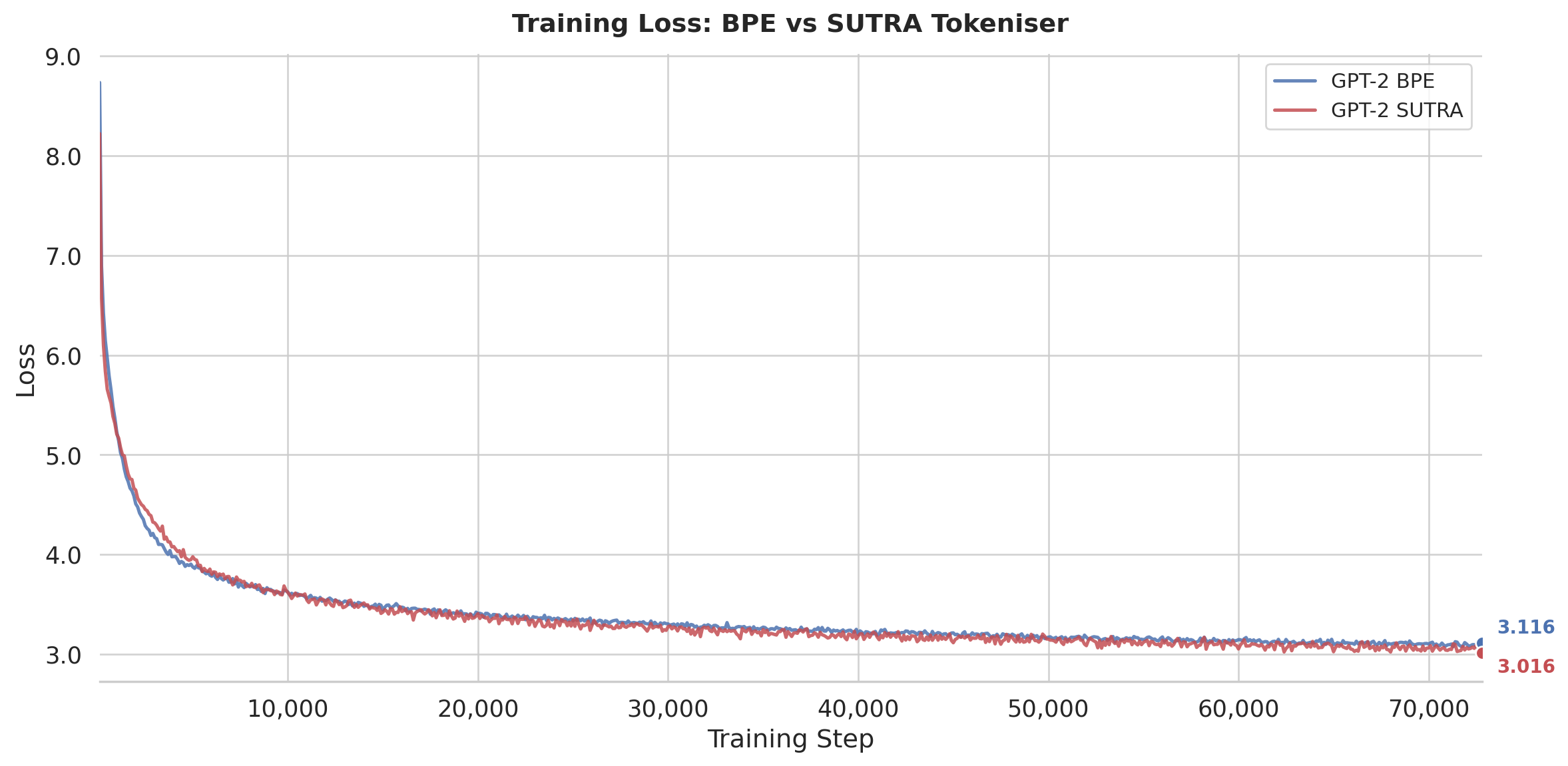}
    \caption{Loss curves for BPE and \textsc{SuTRA}}
    \vspace{-1.5em}
    \label{fig:LM_loss}
\end{figure}
\begin{table}[ht]
\centering
\caption{Downstream performance on Causal Language Modeling (CLM). We report average Negative Log-Likelihood (Avg NLL) and Perplexity (PPL) on out-of-distribution (OOD) corpora. Lower values indicate better compression and predictive efficiency.}
\label{tab:clm_results}
\small
\begin{tabular}{@{}llccc@{}}
\toprule
\textbf{Language} & \textbf{Tokenizer} & \textbf{Vocab Size} & \textbf{Avg NLL $\downarrow$} & \textbf{Perplexity $\downarrow$} \\ \midrule
\multirow{2}{*}{Hindi (HI)} & BPE & 32000 & 3.116 & 18.87 \\
 & \textsc{SuTRA} & 32000 & 3.016 & 16.08 \\
\midrule
\bottomrule
\end{tabular}
\end{table}

\clearpage
\newpage

\newpage

\noindent\section{Extended Analysis of Morphological Robustness Evaluation}

\subsection{Quantifying Morphological Resilience}
Standard tokenizers often lack linguistic awareness, leading to arbitrary root fragmentation when faced with noisy text—a phenomenon we term \textbf{Morphological Shattering}. We center our investigation around the following research question:
\begin{quote}
     To what extent can a semantics-constrained tokenizer maintain the integrity of a language's root-affix structure when faced with orthographic and morphological variations?
\end{quote}
We evaluate tokenizer robustness using an adversarial test set of 10,000 unique words per language, generating variants via synthetic orthographic noise (e.g., character swaps, deletions , substitute , insertions) to simulate real-world typos and morphological shifts. To quantify the tokenizer's susceptibility to catastrophic re-segmentation we compare \textsc{SuTRA} against six baselines using three metrics: \textbf{Jaccard Overlap (Jac. $\uparrow$)}, which calculates the token-level intersection over union to measure structural consistency; \textbf{Root Affected Distance (R.Aff. $\downarrow$)}, which assesses semantic vulnerability by measuring how deeply noise alters the core root's tokenization. As shown in Table~\ref{tab:baseline_comparison}, standard frequency-based methods (\textsc{Bpe}, \textsc{Unigram}) exhibit high Root Affected Distance ($>0.22$), indicating that minor orthographic noise forces them to recalculate boundaries from scratch, thereby corrupting the semantic anchor. In contrast, \textsc{SuTRA} achieves a Root Affected Distance near zero ($\approx 0.04$) . This confirms that enforcing morphological priors successfully insulates the root, preserving linguistic utility even under adversarial conditions.

\begin{table}[htbp]
\centering
\tiny

\resizebox{\linewidth}{!}{%
\begin{tabular}{l|cc|cc|cc}
\toprule
\multirow{2}{*}{\textbf{Tokenizer}} & \multicolumn{2}{c|}{\textbf{Hindi (HI)}} & \multicolumn{2}{c|}{\textbf{Marathi (MR)}} & \multicolumn{2}{c}{\textbf{Gujarati (GU)}} \\
\cmidrule(lr){2-3} \cmidrule(lr){4-5} \cmidrule(lr){6-7}
 & \textbf{Jac.($\uparrow$)} & \textbf{R.Aff.($\downarrow$)} & \textbf{Jac.($\uparrow$)} & \textbf{R.Aff.($\downarrow$)} & \textbf{Jac.($\uparrow$)} & \textbf{R.Aff.($\downarrow$)} \\
\midrule
\textsc{Bpe} & 0.386 & 0.228 & 0.305 & 0.324 & 0.319 & 0.296 \\
\textsc{Unigram} & 0.363 & 0.232 & 0.294 & 0.340 & 0.320 & 0.308 \\
\textsc{WordPiece} & 0.413 & 0.249 & 0.293 & 0.355 & 0.314 & 0.316 \\
\textsc{SentencePiece} & 0.371 & 0.235 & 0.298 & 0.325 & 0.321 & 0.287 \\
\textsc{SuperBpe} & 0.624 & 0.146 & 0.655 & 0.127 & 0.631 & 0.126 \\
\textsc{MorphTok} & 0.801 & 0.067 & 0.823 & 0.055 & 0.792 & 0.051 \\
\midrule
\rowcolor[gray]{0.9} \textbf{\textsc{SuTRA} (Ours)} & \textbf{0.875} & \textbf{0.042} & \textbf{0.885} & \textbf{0.038} & \textbf{0.788} & \textbf{0.040} \\
\bottomrule
\end{tabular}%
}
\setlength{\tabcolsep}{2pt}
\caption{\textbf{Morphological Shattering Comparison.} Results show that \textsc{SuTRA} effectively mitigates root fragmentation.}
\label{tab:baseline_comparison}
\end{table}

To provide a rigorous evaluation of how tokenizers handle the complex morphology of Indic languages, we analyze the relationship between eight distinct metrics. These metrics collectively quantify the \textit{Shattering effect}—the phenomenon where a tokenizer loses semantic consistency when faced with orthographic variations.

\section{Limitations}
\label{sec:limitations}
Vocabulary Scaling. To ensure a controlled comparison across all tokenizers, we maintained a uniform vocabulary size of 32k. While this provides a consistent baseline for evaluating architectural efficiency, the performance of \textsc{SuTRA} under larger vocabulary regimes (e.g., 64k or 128k) remains unexplored. Future benchmarking is required to determine if increased capacity further mitigates Morphological Shattering or if the benefits of semantic constraints saturate at higher scales.

\end{document}